\documentclass[11pt,logo,letterpaper]{berkeley}

\usepackage[utf8]{inputenc}
\usepackage[T1]{fontenc}
\usepackage{tgpagella}
\usepackage{microtype}

\usepackage{amsmath,amsthm,amsfonts}
\usepackage{mathrsfs}
\usepackage{bm}
\usepackage{latexsym}
\usepackage{bbding}
\usepackage{pifont}

\usepackage{marvosym}

\usepackage{xcolor}
\usepackage{graphicx}
\usepackage{subcaption}
\usepackage{tabularx}
\usepackage{adjustbox}
\usepackage{wrapfig}
\usepackage{float}
\usepackage{tikz}
\usetikzlibrary{shapes.geometric, arrows.meta, positioning, fit}
\usepackage[edges]{forest}
\usepackage{rotating}
\usepackage{eso-pic}
\usepackage{booktabs}
\usepackage{multirow}
\usepackage{array}
\usepackage{tabularx}
\usepackage{colortbl}
\usepackage{threeparttable}
\usepackage{longtable}
\usepackage{lscape}
\usepackage{makecell}

\usepackage{amssymb}

\definecolor{YaleBlue}{HTML}{2A5487}
\definecolor{CalGoldHex}{HTML}{FDF7F2}
\definecolor{TikTokPink}{HTML}{FF2B54}
\definecolor{IronGrey}{HTML}{6D6E71}
\definecolor{LinkPurple}{HTML}{FF2B54}
\definecolor{pipeline-orange}{RGB}{241,146,83}	
\definecolor{tree-pink}{RGB}{255,232,242}
\definecolor{tree-cyan}{RGB}{199,241,240}
\definecolor{tree-red}{RGB}{255,215,214}
\definecolor{tree-purple}{RGB}{230,231,255}
\definecolor{tree-green}{RGB}{192,242,213}
\definecolor{tree-yellow}{RGB}{255,242,230}
\definecolor{tree-blue}{RGB}{223,244,255}

\definecolor{box-pink}{RGB}{255,72,162}
\definecolor{box-cyan}{RGB}{29,234,221}
\definecolor{box-red}{RGB}{255,1,25}
\definecolor{box-purple}{RGB}{61,31,255}
\definecolor{box-green}{RGB}{71,212,90}
\definecolor{box-yellow}{RGB}{255,146,1}
\definecolor{box-blue}{RGB}{5,188,248}

\definecolor{hidden-draw}{RGB}{0,0,0}
\definecolor{hidden-pink}{RGB}{255,245,247}
\definecolor{hidden-red}{RGB}{205,44,36}
\definecolor{hidden-blue}{RGB}{194,232,247}
\definecolor{hidden-orange}{RGB}{243,202,120}
\definecolor{hidden-green}{RGB}{34,139,34}
\definecolor{hidden-black}{RGB}{20,68,106}
\definecolor{hidden-yellow}{RGB}{255,248,203}
\definecolor{purple}{RGB}{144,153,196}
\definecolor{yellow}{RGB}{255,228,123}
\definecolor{tkcolor}{RGB}{224,223,255}
\definecolor{level0}{rgb}{0.67, 0.88, 0.69}
\definecolor{level1}{rgb}{0.98, 0.92, 0.84}
\definecolor{level2}{rgb}{0.8, 0.8, 1.0}
\definecolor{level3}{rgb}{1.0, 0.71, 0.76}
\definecolor{level4}{rgb}{0.49, 0.99, 0.0}
\definecolor{level5}{rgb}{0.87, 0.63, 0.87}
\definecolor{darkblue}{rgb}{0, 0.40, 0.75}
\definecolor{mygreen}{RGB}{144, 238, 144}
\definecolor{darkmygreen}{RGB}{0, 100, 0}

\usepackage{caption}
\usepackage[square,sort,comma,numbers]{natbib}
\usepackage{enumitem}
\setlist{
  itemsep=2pt,
  parsep=1pt,
  topsep=0pt,
  partopsep=0pt,
  leftmargin=*
}
\setlist[itemize]{itemsep=2pt, topsep=3pt, parsep=1pt}
\setlist[enumerate]{itemsep=3pt, topsep=4pt, parsep=1pt}
\setlist[description]{itemsep=3pt, topsep=4pt, parsep=1pt, style=nextline}
\usepackage{multicol}
\usepackage{titletoc}
\usepackage{titlesec}
\usepackage{algorithm,algpseudocode}
\usepackage{tcolorbox}
\usepackage{fontawesome}

\usepackage{url}
\usepackage{hyperref}
\hypersetup{colorlinks=true, linkcolor=YaleBlue}
\usepackage{cleveref}

\tikzstyle{my-box}=[
rectangle,
draw=black,
rounded corners,
text opacity=1,
minimum height=1.5em,
minimum width=5em,
inner sep=2pt,
align=left,
fill opacity=.5,
]

\tikzstyle{section_2}=[my-box, fill=tree-pink]
\tikzstyle{section_3}=[my-box, fill=tree-blue]
\tikzstyle{section_4}=[my-box, fill=tree-cyan]
\tikzstyle{section_5}=[my-box, fill=tree-green]
\tikzstyle{section_6}=[my-box, fill=tree-purple]
\tikzstyle{appendix}=[my-box, fill=tree-yellow]

\tikzstyle{leaf}=[my-box, minimum height=1.5em, fill=tree-pink, text=black, align=left, font=\normalsize, inner xsep=5pt, inner ysep=4pt, text width=45em]
\tikzstyle{leaf2}=[my-box, minimum height=1.5em, fill=tree-cyan, text=black, align=left, font=\normalsize, inner xsep=5pt, inner ysep=4pt]
\tikzstyle{leaf3}=[my-box, minimum height=1.5em, fill=tree-red, text=black, align=left, font=\normalsize, inner xsep=5pt, inner ysep=4pt]
\tikzstyle{leaf4}=[my-box, minimum height=1.5em, fill=tree-purple, text=black, align=left, font=\normalsize, inner xsep=5pt, inner ysep=4pt]

\newcommand{\github}{\raisebox{-1.5pt}{\includegraphics[height=1.05em]{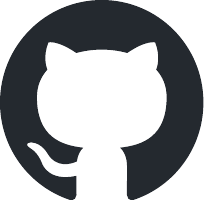}}}

\newcommand{\MYhref}[3][YaleBlue]{\href{#2}{\color{#1}{#3}}}%

\newlength{\sectionbeforeskip}
\newlength{\sectionafterskip}
\newlength{\subsectionbeforeskip}
\newlength{\subsectionafterskip}
\newlength{\subsubsectionbeforeskip}
\newlength{\subsubsectionafterskip}
\newlength{\openingafterspace}
\newlength{\epigraphafterspace}
\newlength{\leadparagraphskip}
\newlength{\guidingquestionaboveskip}
\newlength{\guidingquestionbelowskip}
\newlength{\sectionsubtitleaboveskip}
\newlength{\sectionsubtitlebelowskip}
\newlength{\remarkboxskip}
\newlength{\tocheaderskip}
\newlength{\tocrulesep}
\newlength{\tocbodyskip}

\titlespacing*{\section}{0pt}{\sectionbeforeskip}{\sectionafterskip}
\titlespacing*{\subsection}{0pt}{\subsectionbeforeskip}{\subsectionafterskip}
\titlespacing*{\subsubsection}{0pt}{\subsubsectionbeforeskip}{\subsubsectionafterskip}

\newcommand{\openingstatement}[1]{%
  \begin{quote}
    \color{YaleBlue}\itshape #1
  \end{quote}
  \vspace{\openingafterspace}
}

\newcommand{\chapterepigraph}[3]{%
  \begin{quote}
    \color{YaleBlue}\itshape
    ``#1''\\[0.8ex]
    {\normalfont\small\hspace*{\fill}--- \textsc{#2}%
    \if\relax\detokenize{#3}\relax\else\\
      \hspace*{\fill}\footnotesize\textit{#3}%
    \fi}
  \end{quote}
  \vspace{\epigraphafterspace}
}

\newcommand{\leadparagraph}[1]{%
  \par\vspace{\leadparagraphskip}\noindent\textbf{#1}\enspace
}

\newcommand{\labeledlead}[2]{%
  \leadparagraph{#1. #2}
}

\newcommand{\sectionsubtitle}[1]{%
  \par\vspace{\sectionsubtitleaboveskip}%
  \noindent{\color{IronGrey}\itshape #1}%
  \par\vspace{\sectionsubtitlebelowskip}
}

\newcommand{\rendercondensedtoc}{%
  \addtocontents{toc}{\protect\setcounter{tocdepth}{2}}%
  \vspace*{\tocheaderskip}%
  \startcontents[sections]\vbox{\sc Table of Contents}%
  \vspace{\tocrulesep}%
  \hrule height .8pt
  \vspace{-2mm}
  {\setlength{\baselineskip}{11pt}%
  \setlength{\parskip}{3pt}%
  \printcontents[sections]{l}{1}{\setcounter{tocdepth}{2}}}%
  \vspace{\tocrulesep}%
  \hrule height .8pt
  \vspace*{\tocbodyskip}%
}

\tcbset{
  takeawaysbox/.style={
    colback=CalGoldHex,
    colframe=YaleBlue,
    coltext=IronGrey,
    fontupper=\footnotesize,
    width=\linewidth,
    arc=2mm,
    boxrule=0.2mm,
    top=4mm,
    bottom=3mm,
    left=5mm,
    right=5mm,
    before skip=\remarkboxskip,
    after skip=\remarkboxskip,
  }
}

\newtcolorbox[auto counter]{remarkboxenv}[2][]{
  takeawaysbox,
  colbacktitle=YaleBlue,
  coltitle=white,
  fonttitle=\bfseries,
  title={Remark \thetcbcounter: #2},
  #1
}

\newcommand{\xmark}{\ding{55}}

\tcbuselibrary{breakable}
\newcommand{\remarkbox}[1]{
\begin{tcolorbox}[
    colback=gray!10, 
    colframe=gray!50, 
    coltitle=black, 
    width=\textwidth, 
    boxrule=0.5pt, 
    arc=1mm, 
    boxsep=1mm,
    left=3mm,
    right=3mm,
    top=3mm,
    bottom=3mm,
    before skip=4mm,
    after skip=4mm,
    breakable
]
\textbf{Remarks}. {#1}
\end{tcolorbox}
}

\title{Token Economics for LLM Agents: A Dual-View Study from Computing and Economics}

\runningtitle{Token Economics for LLM Agents: A Dual-View Study from Computing and Economics}

\author{\textbf{Yuxi Chen}\textsuperscript{\rm 1,2 $\ast$}
\hfill \textbf{Junming Chen}\textsuperscript{\rm 1 $\ast$}
\hfill \textbf{Chenyu He}\textsuperscript{\rm 1 $\ast$}
\hfill \textbf{Yiwei Li}\textsuperscript{\rm 2 $\ast$}
\hfill \textbf{Yicheng Ji}\textsuperscript{\rm 1 $\ast$}
\hfill \textbf{Yifan Wu}\textsuperscript{\rm 1,3 $\ast$} \\ 
\textbf{Dingyu Yang}\textsuperscript{\rm 1,3} 
\hfill \textbf{Lansong Diao}\textsuperscript{\rm 4}
\hfill \textbf{Lidan Shou}\textsuperscript{\rm 1,3}
\hfill \textbf{Hongliang Zhang}\textsuperscript{\rm 2 \Letter}
\hfill \textbf{Huan Li}\textsuperscript{\rm 1,3 \Letter}
\hfill \textbf{Gang Chen}\textsuperscript{\rm 1}\\ 
\vspace{2mm}
\textsuperscript{\rm 1} College of Computer Science and Technology, Zhejiang University \\
\textsuperscript{\rm 2} School of Economics, Zhejiang University \enskip
\textsuperscript{\rm 3} The State Key Laboratory of Blockchain and Data Security, Zhejiang University \enskip
\textsuperscript{\rm 4} Alibaba Cloud
}

\begin{document}

\AddToShipoutPictureBG*{
  \AtPageUpperLeft{
    \put(\LenToUnit{2.0cm},\LenToUnit{-2.2cm}){ 
      \includegraphics[width=9.5cm, height=9.5cm, keepaspectratio]{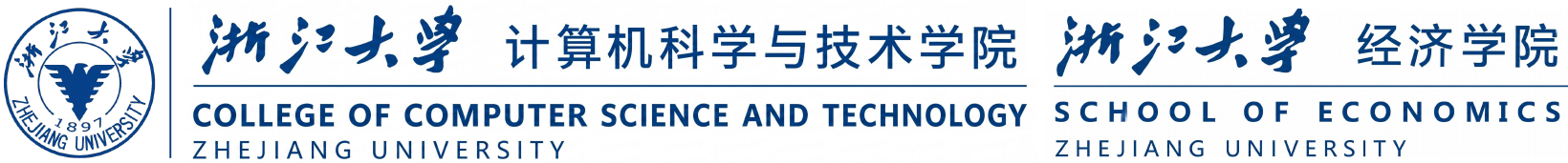}
    }
  }
}
\AddToShipoutPictureBG*{
  \AtPageUpperLeft{
    \put(\LenToUnit{11.6cm},\LenToUnit{-2.15cm}){ 
      \includegraphics[width=4.8cm, height=4.8cm, keepaspectratio]{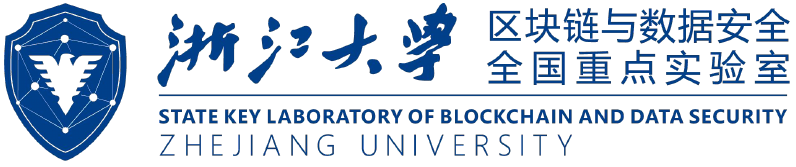}
    }
  }
}
\AddToShipoutPictureBG*{
  \AtPageUpperLeft{
    \put(\LenToUnit{16.6cm},\LenToUnit{-2.0cm}){ 
      \includegraphics[width=2.6cm, height=2.6cm, keepaspectratio]{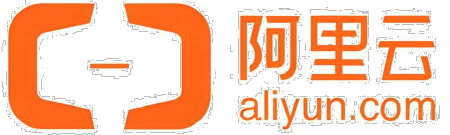}
    }
  }
}

\begingroup
\renewcommand{\thefootnote}{}
\footnotetext{\textsuperscript{$\ast$}Equal Contribution. \quad \textsuperscript{\rm \Letter}Corresponding Author.}
\endgroup

\begin{abstract}
As LLM agents evolve, tokens have emerged as the core economic primitives of Agentic AI. However, their exponential consumption introduces severe computational, collaborative, and security bottlenecks. Current surveys remain fragmented across system optimization, architecture design, and trust, lacking a unified framework to evaluate the fundamental trade-off between output quality and economic cost.
To bridge this gap, this survey presents the first comprehensive survey of Token Economics. By unifying computer science and economics, we conceptualize tokens as production factors, exchange mediums, and units of account. We synthesize existing literature across a four-dimensional taxonomy:
(1) Micro-level (Single Agent): Optimizing budget-constrained factor substitution via neoclassical firm theory.
(2) Meso-level (Multi-Agent Systems): Minimizing collaboration friction using transaction cost and principal-agent theories.
(3) Macro-level (Agent Ecosystems): Addressing congestion externalities and pricing via mechanism design.
(4) Security: Internalizing adversarial threats as endogenous economic constraints.
Finally, we outline frontier directions, including differentiable token budgets and dynamic markets, to lay the theoretical foundation for scalable next-generation agent systems.

\smallskip
\centering
\github{} \textbf{GitHub}: \MYhref{https://github.com/SuDIS-ZJU/Token-Economics}{\textit{\textbf{https://github.com/SuDIS-ZJU/Token-Economics}}}

\bigskip
\par\vspace{0.2em}
\centering
\includegraphics[width=1.0\linewidth]{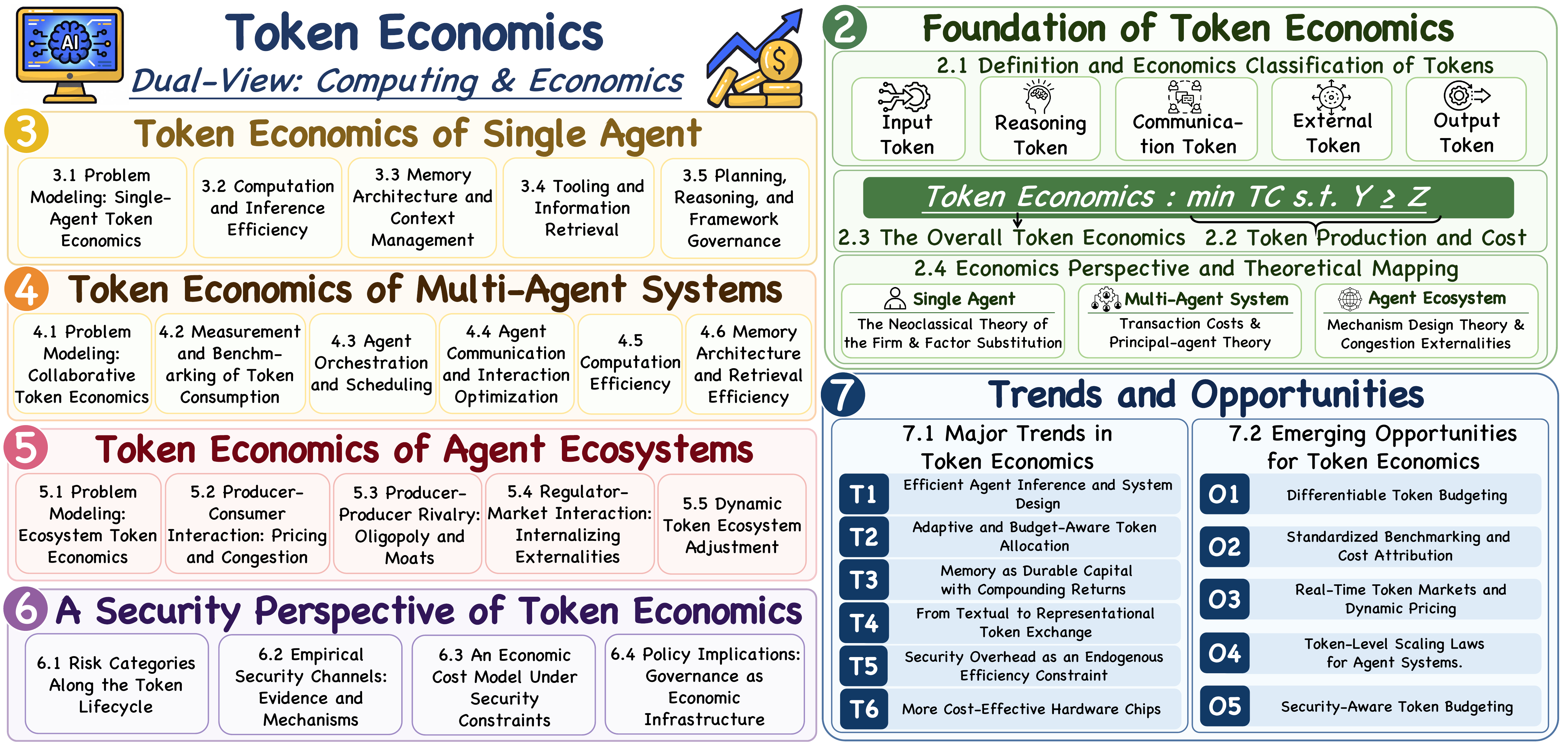}
\par\vspace{0.2em}
{\small Overview of the dual-view survey on token economics.}
\label{fig:paper_overview}

\end{abstract}

\maketitle

\clearpage
\vspace*{-1cm}
\rendercondensedtoc

\openingstatement{Countless LLM tokens were consumed to synthesize this survey, the product of an intensive dialogue between machine computation and human insight. This practical reality underscores our message. We present this work with the firm conviction that tokens have evolved far beyond simple data units; by bridging economic theory and resource-efficient system design, we reveal them as the foundational currency of our intelligence-driven future---\textbf{Token Economics}.}

\section{Introduction}
\label{sec:intro}

Historically, major technological epochs have been defined by shifts in their foundational economic primitives. The kilowatt-hour (kWh) galvanized the Industrial Age, and network bandwidth (GB) underpinned the Information Age. Today, the ``\textbf{\emph{token}}'' is powering the Intelligence Age, the era of generative AI and large language model (LLM) agents, by serving as the universal substrate of digital creation. Every multimodal interaction, whether text, vision, or sound, is ultimately distilled into token flows; through those flows, human cognition is translated into machine execution. In this new paradigm, the token no longer functions merely as a technical unit of computation. It has become the economic primitive of agentic AI~\cite{DBLP:journals/corr/abs-2505-18227}: the fundamental unit by which intelligence is produced and measured, and the practical currency by which it is exchanged. In this role, it follows the iron logic of any foundational resource: as the economy built atop it expands, so too does the demand for the resource itself. The token was thus destined to be consumed at a scale that defies linear extrapolation.

\begin{figure}[!htbp]
  \centering
  \includegraphics[width=0.98\linewidth, height=0.35\textheight, keepaspectratio]{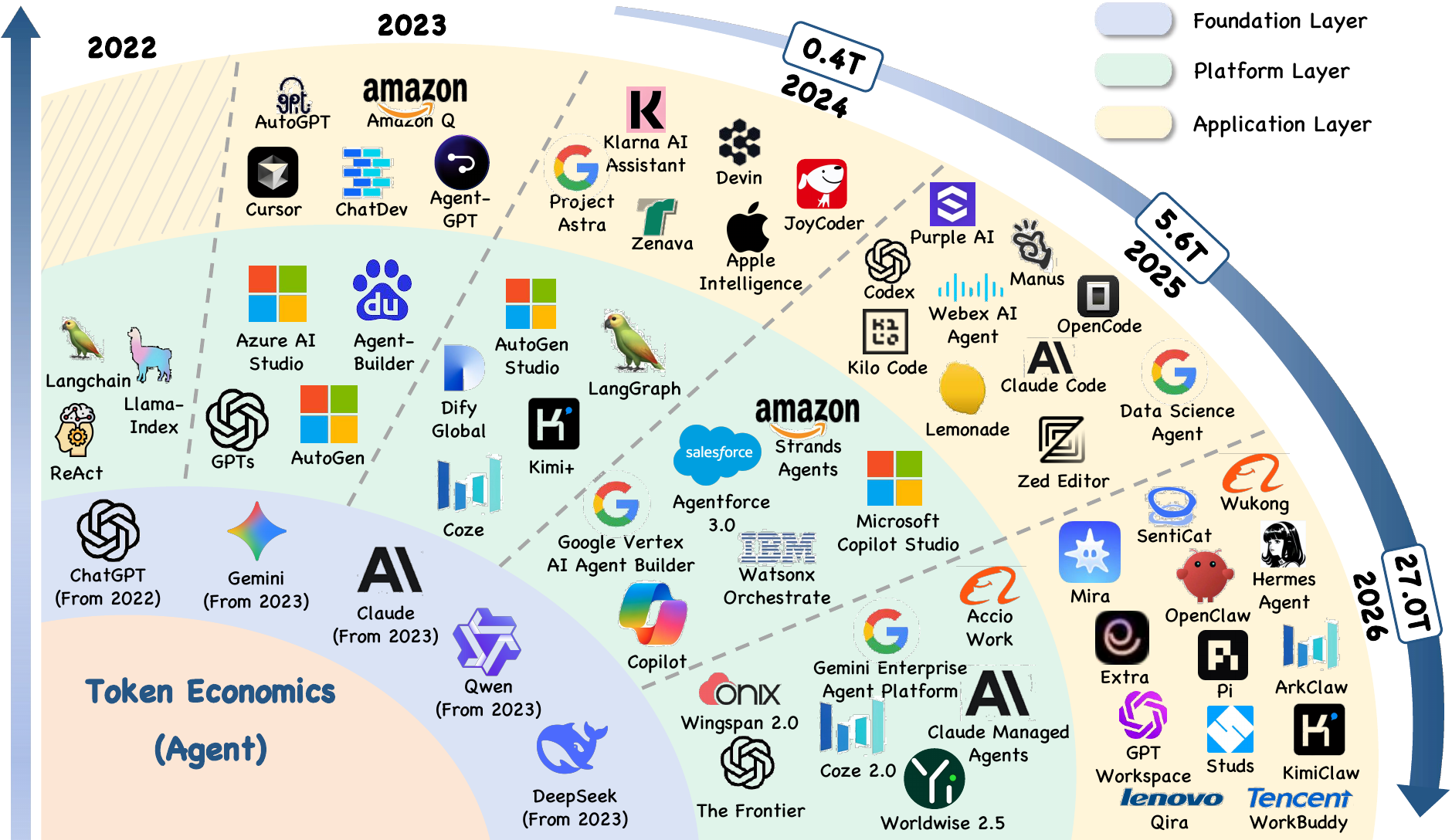}
  \caption{Evolution (Year 2022--2026) of the agent technology stack across foundation, platform, and application layers. \textbf{\textcolor{YaleBlue}{Values in boxes}} denote the weekly usage of models across OpenRouter at key milestones (Dec. 2024, Dec. 2025, Mar. 2026).}
  \label{fig:token_num}
\end{figure}

This trend is already unfolding, and nowhere is it more visible than in the rise of agentic AI~\cite{zhong2024memorybank,yue2025masrouter,bian2026tokendance}. Unlike traditional single-pass LLM inference, agent workflows operate through iterative loops of reasoning, tool use, and self-correction, each cycle consuming tokens as a direct input to cognition. Moreover, because agent workflows are inherently far more token-intensive than conventional LLM calls, their proliferation has driven an exponential surge in consumption. As more and more agent platforms and end-user applications emerge (\Cref{fig:token_num}), weekly token processing volume on the OpenRouter platform skyrocketed from 0.4 trillion in December 2024 to 27.0 trillion by March 2026, a nearly 68-fold increase in just 15 months~\footnote{Data sourced from OpenRouter: \url{https://openrouter.ai/rankings}}. LLM agents, and the token flows that sustain them, are no longer confined to experimental settings; they are now embedded in high-stakes domains such as finance~\cite {li2025investorbench, barry2025graphrag, li2026time}, law~\cite{akarajaradwong2025nitibench, li2025legalagentbench, li2025lexrag}, and healthcare~\cite{cheng2026novo, zhu2025ask, li2026tumorchain}. This trajectory has transformed token consumption from a technical detail into a systemic pressure point.

This pressure is now materializing as a tangible supply-demand crisis. The unchecked expansion of token consumption is driving a synchronous surge in computational resource demands across individuals, enterprises, and society. The International Energy Agency projects that global data center electricity usage will double by 2030~\cite{IEA2025EnergyAI}. This growing imbalance has pushed the industry to a critical inflection point, forcing a strategic pivot \textbf{\emph{from merely scaling compute to optimizing token efficiency}}.
Against this backdrop, the industry's shift toward conceptualizing AI data centers as ``AI factories''~\footnote{NVIDIA CEO Jensen Huang prominently articulated related vision during the NVIDIA GTC Keynote 2026. Official source: \url{https://www.youtube.com/watch?v=jw_o0xr8MWU} (Timestamp: 01:06:46)}, a conceptual shift that has catalyzed the formalization of \textbf{Token Economics}.

This view reframes the core proposition straightforwardly: how can LLM-agent systems generate high-quality tokens (\textbf{\emph{superior performance, lower cost, and enhanced security}}) under strict computational budgets and latency constraints? Under this perspective, inference acceleration and algorithmic optimization are no longer mere engineering choices; they are economic imperatives that shape the sustainability of the agent ecosystem.

\begin{figure}[!htbp]
  \centering
  \includegraphics[width=1.0\linewidth, height=0.35\textheight, keepaspectratio]{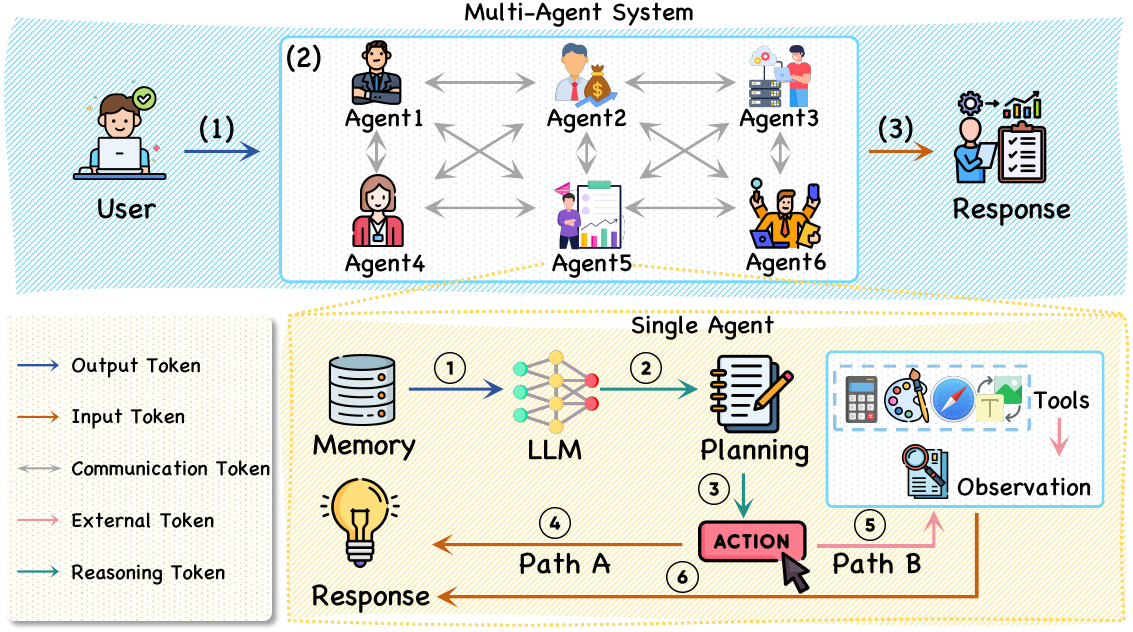}
  \caption{The top layer illustrates the Multi-Agent System (MAS) coordinating inter-agent synchronization (1)--(3) via Communication Tokens. The bottom layer details the Single-Agent's internal ``Memory-Planning-Action'' micro-loop (\ding{172}-\ding{177}).}
  \label{fig:pipeline}
\end{figure}

The proliferation of agent architectures amplifies the tension between computation and economics, reshaping frontier research~\cite{ren2026transcending}. Unlike the linear token consumption in conventional LLMs~\cite{sharma2025ttd}, modern agent workflows (\Cref{fig:pipeline}) are highly iterative. At the multi-agent system (MAS) level, token flow transitions from (1) Input Tokens, through (2) inter-agent Communication Tokens, to (3) Output Tokens.
To drive this macro-flow, individual agents execute a ``Memory-Planning-Action'' micro-loop.
In particular, an agent (see the lower part of \Cref{fig:pipeline}) loads context (\ding{172}), generates Reasoning Tokens to plan (\ding{173}), and triggers actions (\ding{174}). It then updates its state via either internal reasoning (\ding{175} Path A) or external tool observations (\ding{176}\ding{177} Path B). The resulting token is fed back into the MAS layer (2) to sustain ongoing negotiation.
Crucially, completing complex tasks requires repeated reflection, retrieval, and multi-agent synchronization~\cite{niu2025flow}. This shift from isolated inference to organizational coordination introduces substantial internal transaction costs and redundant overheads~\cite{leong2025amas, zhang2024agentprune} that cannot be captured by any single technical dimension. 

The academic community has already developed multidimensional research trajectories around token economics. These include inference acceleration mechanisms, toolchain invocation optimization, and agent memory systems. However, existing survey literature remains heavily compartmentalized into isolated technical silos. Most prior surveys fall into three broad camps (see~\Cref{tab:survey_comparison}):

\begin{itemize}
    \item \textbf{Agent Architecture}: Works by Li et al.~\cite{li2024survey} and Xi et al.~\cite{xi2025rise} provide systematic reviews of multi-agent workflows, construction frameworks, and the evolution of agent-based social simulations.
    \item \textbf{System Optimization}: Miao et al.~\cite{miao2025towards} and Xu et al.~\cite{xu2025resource} analyze resource-efficient serving methodologies and system-level optimizations for deploying large foundation models from an MLSys perspective.
    \item \textbf{Trust and Security}: Deng et al.~\cite{deng2025ai} and He et al.~\cite{he2025emerged} categorize emerging privacy risks, adversarial threats, and defensive strategies across diverse agent operational environments.
\end{itemize}

\begin{table}[!htbp]
\centering
\caption{Overview of prior related surveys.}
\label{tab:survey_comparison}
\renewcommand{\arraystretch}{0.9}
\resizebox{\linewidth}{!}{ 
\begin{tabular}{lclccccc}
\toprule
\multirow{2}{*}{\textbf{Reference}} & \multirow{2}{*}{\textbf{Year}} & \multirow{2}{*}{\textbf{Primary Focus}} & \multicolumn{5}{c}{\textbf{Research Dimensions}} \\
\cmidrule(lr){4-8}
& & & \makecell{\textbf{Capability, Reasoning}}
    & \makecell{\textbf{System Resource, Utilization}} 
    & \makecell{\textbf{Interaction, Communication Friction}} 
    & \makecell{\textbf{Security, Robust, Attrition}} 
    & \makecell{\textbf{Economics}} \\
\midrule
Li et al.~\cite{li2024survey}     & 2024 & Agent Architecture    & \ding{51} & \ding{55} & \ding{51} & \ding{55} & \ding{55} \\
Xi et al.~\cite{xi2025rise}       & 2025 & Agent Architecture    & \ding{51} & \ding{55} & \ding{51} & \ding{55} & \ding{55} \\
Miao et al.~\cite{miao2025towards}& 2025 & System Optimization & \ding{55} & \ding{51} & \ding{55} & \ding{55} & \ding{55} \\
Xu et al.~\cite{xu2025resource}   & 2025 & System Optimization & \ding{55} & \ding{51} & \ding{55} & \ding{55} & \ding{55} \\
Deng et al.~\cite{deng2025ai}     & 2025 & Trust and Security   & \ding{55} & \ding{55} & \ding{51} & \ding{51} & \ding{55} \\
He et al.~\cite{he2025emerged}    & 2025 & Trust and Security   & \ding{51} & \ding{55} & \ding{51} & \ding{51} & \ding{55} \\
\midrule
\rowcolor{blue!10}
\textbf{Ours}                     & \textbf{/} & \textbf{Token Economics}  & \ding{51} & \ding{51} & \ding{51} & \ding{51} & \ding{51} \\
\bottomrule
\end{tabular}
}
\end{table}

This fragmentation creates a central limitation: there is still \textbf{no unified language for measuring the systemic trade-off between algorithmic capability and coordination overhead}. 
Because existing surveys do not treat the token as a fundamental economic primitive---and, more specifically, as a factor of production, a medium of exchange, and a unit of account---they cannot fully explain why locally optimal engineering choices often trigger global diseconomies of scale in complex agent workflows. Within isolated research silos, improving one dimension often imposes hidden costs on another. For example, aggressively maximizing system throughput may compromise reasoning quality, while rigid security defenses may exacerbate token-economic attrition. Without a unified economic lens, the true \textbf{Product-Cost Pareto frontier} remains difficult to characterize.

To move beyond these fragmented heuristics, this survey presents a holistic synthesis of the complete token lifecycle. Through a \textbf{Dual-View} perspective, we connect computational systems with economic theory, bringing algorithmic logic, system utilization, interactive friction, and security overhead into one coherent analytical framework. We anchor this synthesis in the evolution of agent architectures and use economics not merely as a metaphor, but as a structural lens. As AI systems scale, the nature of token friction changes. At the micro-boundary of a Single-Agent (\Cref{sec:single_agent}), the system resembles a solitary firm balancing context retrieval against reasoning depth. In MAS (\Cref{sec:multi_agent}), the architecture resembles a corporate hierarchy, and the bottleneck shifts toward transaction costs, especially the communication tokens required for state synchronization and conflict resolution. At the Ecosystem scale (\Cref{sec:ecosystem}), the dynamics resemble an open market constrained by macro-externalities such as adversarial security attrition and multi-tenant congestion. Finally, to ensure agentic systems robustly achieve the Pareto frontier amidst realistic adversarial environments, we conceptualize agent security (\Cref{sec:eco_security}) as a critical external shock.

By rigorously tracing this organizational trajectory, our mapping bridges the gap between abstract economic theory and empirical computer systems, and organizes Token Economics into a coherent field-level blueprint. Specifically, the core contributions of this survey are summarized as follows:

\begin{itemize}
    \item \textbf{A unifying dual-view framework.} We establish a formal conceptualization of \textit{Token Economics}. By linking computational systems with economic theory, we conceptualize tokens simultaneously as factors of production, media of exchange, and units of account, and redefine LLM agent inference as a constrained resource-allocation problem under one shared economic language.
    
    \item \textbf{Systematic categorization across architectural scales.} Guided by economics, we taxonomize the fragmented state of the art from theoretical foundations to single-agent optimization, multi-agent coordination, ecosystem-level allocation, and security economics. This structure clarifies the distinct technical and economic bottlenecks that emerge at each structural boundary.
    
    \item \textbf{Internalizing security and charting a future roadmap.} We move beyond traditional capability metrics by reframing adversarial vulnerabilities and alignment mechanisms not merely as isolated compliance constraints, but as endogenous sources of token-economic attrition. Building on this view, we outline a research roadmap spanning differentiable token budgeting, memory capital accumulation, and dynamic token markets.
\end{itemize}

\begin{figure}[htbp]
\vspace{-2mm}
\centering
\resizebox{0.95\textwidth}{!}{
	\begin{forest}
		forked edges,
		for tree={
			grow=east,
			reversed=true,
			anchor=base west,
			parent anchor=east,
			child anchor=west,
			base=left,
			font=\large,
			rectangle,
			draw=black,
			rounded corners,
			align=left,
			minimum width=4em,
			edge+={darkgray, line width=1pt},
			s sep=10pt,
			inner xsep=2pt,
			inner ysep=4pt,
			line width=1.1pt,
			ver/.style={rotate=90, child anchor=north, parent anchor=south, anchor=center},
		},
		where level=1{text width=10em,font=\normalsize,}{},
        where level=2{text width=15em,font=\normalsize,}{},
        where level=3{text width=15em,font=\normalsize,}{},
        where level=4{text width=45em,font=\normalsize,}{},
[\ \ \ \ \ \ \ \ {Token Economics for LLM Agents: A Dual-View Study from Computing and Economics}, text width=50em, ver
	[\ \ \ \ \ \ \ \ Foundation of Token \\ \ \ \ \ \ \ \ \ \ \ \ Economics~(\S\ref{sec:concepts}), text width=12em, section_2, draw=box-pink,  text width=18em
            [\ \ \textbf{Definition and Economics Classification of Tokens}~(\S\ref{sec:concepts_classification}), section_2, draw=box-pink,  text width=65em
            ]
            [\textbf{\ \ \ Token Production and Cost}~(\S\ref{sec:concepts_production}), section_2, draw=box-pink,  text width=18em
                [\ \ Production Function~\citep{behrman2024tutoring}{.},
                leaf, section_2, draw=tree-pink
    		  ]
                [\ \ Cost Function~\citep{raval2023testing}{.},
                leaf, section_2, draw=tree-pink
    		  ]
            ]
            [\ \ \textbf{The Overall Token Economics}~(\S\ref{sec:concepts_te}), section_2, draw=box-pink,  text width=65em
            ]
		[\textbf{\ \ \ \ \ \ \ \ \ \ Economics Perspective} \\ \textbf{\ \ \ \ \ and Theoretical Mapping}~(\S\ref{sec:concepts_organizational}), section_2, draw=box-pink,  text width=18em
            [\ \ Single Agent~\citep{schick2023toolformer} $\rightarrow$ The Neoclassical Theory of the Firm~\citep{chodorow2025neoclassical}{,}
            Factor Substitution~\citep{hubmer2026not}{.}
            , leaf, section_2, draw=tree-pink]
            [\ \ Multi-Agent System~\citep{yue2025masrouter} $\rightarrow$ Transaction Costs~\citep{patil2024firm}{,}
            Principal-agent Theory~\citep{lavi2022principal}{.}
            , leaf, section_2, draw=tree-pink]
            [\ \ Agent Ecosystem~\citep{basu2023stablefees} $\rightarrow$ Mechanism Design Theory~\citep{pycia2023theory}{,}
            Congestion Externalities~\citep{ershov2024variety}{.}
            , leaf, section_2, draw=tree-pink]
		]
	]
    [\ \ \ \ \ \ \ \ Token Economics of \\ \ \ \ \ \ \ the Single Agent~(\S\ref{sec:single_agent}), section_3, draw=box-blue, text width=15em
    [\ \ \textbf{Problem Modeling: Single-Agent Token Economics} (\S\ref{subsec:single_problem}), section_3, draw=box-blue,  text width=65em
    ]
    [\textbf{\ \ \ \ \ \ \ \ \ \ \ \ \ \ Computation and} \\ \textbf{\ \ \ \ \ \ \ \ \ \ \ Inference Efficiency} (\S\ref{subsec:single_compute}), section_3, draw=box-blue,  text width=18em
        [\ \ Token Density~\citep{bengio2003neural,mikolov2013efficient,pennington2014glove,sennrich2016neural,kudo2018sentencepiece,oord2017neural,razavi2019generating}{,} 
        Token Quantity~\citep{cheng2024compressed,shen2025codi,wang2025r1compress,xia2025tokenskip,hao2025coconut,yang2026dynamicearlyexit,jiang2025flashthink}{,} \\
        \ \ Per-Token Computation and Memory Cost~\citep{dao2022flashattention,zaheer2020big,katharopoulos2020transformers,li2024snapkv,zhang2023h2o,frantar2023optq,sanh2020movement}{,} \\
        \ \ MoE architectures~\citep{fedus2022switch,zoph2022stmoe}{,}
        Speculative decoding~\citep{leviathan2023speculative,zhang2024draft}{.}
	, leaf2, text width=45em, section_3, draw=tree-blue]
    ]
    [\textbf{\ \ \ \ \ \ \ \ \ Memory Architecture and} \\ \textbf{\ \ \ \ \ \ \ \ Context Management} (\S\ref{subsec:single_memory}), section_3, draw=box-blue,  text width=18em
        [\ \ Context Window as Working Memory~\citep{jiang2023llmlingua,jiang2024longllmlingua,mu2023learning,chevalier2023autocompressors,li2023selective}{,} \\
        \ \ Long-Term and Episodic Memory~\citep{packer2024memgpt,park2023generative,shinn2023reflexion,zhong2024memorybank,xu2025amem,chhikara2025mem0}{.}
	, leaf2, text width=45em, section_3, draw=tree-blue]
    ]
    [\textbf{\ \ \ \ \ \ \ \ \ \ \ \ \ \ \ \ \ \ Tooling and} \\ \textbf{\ \ \ \ \ \ \ \ \ Information Retrieval} (\S\ref{subsec:single_tool}), section_3, draw=box-blue,  text width=18em
        [\ \ Tool Calling and Function Invocation~\citep{schick2023toolformer,patil2023gorilla,qin2024toolllm,du2024anytool,toolrl2024,hou2025model}{,} \\
        \ \ Retrieval-Augmented Generation~\citep{selfrag2024,yan2024corrective,jeong2024adaptive,edge2024graphrag,sarthi2024raptor,gao2023hyde,du2026arag}{.}
	, leaf2, text width=45em, section_3, draw=tree-blue]
    ]
    [\textbf{\ \ \ \ \ \ \ \ \ \ Planning, Reasoning, and} \\ \ \textbf{\ \ \ \ \ \ \ Framework Governance} (\S\ref{subsec:single_planning}), section_3, draw=box-blue,  text width=18em
        [\ \ The Reasoning Investment Spectrum~\citep{cot2022,yao2023react,yao2023tree,besta2024graph,zhou2024lats,li2026bavt,wang2023plan}{,}\\ 
        \ \ Agent Frameworks and Organizational Design\citep{sumers2024cognitive,wang2023voyager,yang2024sweagent,he2026harness}{.}
	, leaf2, text width=45em, section_3, draw=tree-blue]
    ]
]
[\ \ \ \ \ \ \ \ Token Economics in \\ \ \ \ \ \ Multi-Agent Systems~(\S\ref{sec:multi_agent}), text width=10em, section_4, draw=box-cyan, text width=15em
    [\textbf{\ \ Problem Modeling: Collaborative Token Economics}~(\S\ref{subsec:multi_problem}), section_4, draw=box-cyan,  text width=65em]
    [\textbf{\ \ \ \ \ \ Measurement and Benchmarking} \\ \textbf{\ \ \ \ \ \ \ \ of Token Consumption}~(\S\ref{subsec:mas_benchmark}), section_4, draw=box-cyan, text width=18em
        [\ \ Token Distribution Analysis and Taxonomies~\citep{wang2025agenttaxo,salim2026tokenomics,bai2026howdo,zhu2025multiagentbench}{,} \\
        \ \ Scaling Laws and Cross-Framework Evaluation~\citep{yin2025comprehensive,kim2025towards}{.}
        , leaf3, text width=45em, section_4, draw=tree-cyan]
        ]
    [\textbf{\ \ \ \ \ \ \ \ \ \ Agent Orchestration and} \\ \textbf{\ \ \ \ \ \ \ \ \ \ \ \ \ \ \ Scheduling}~(\S\ref{subsec:mas_orchestration}), section_4, draw=box-cyan,  text width=18em
        [\ \ Communication Graph Pruning and Agent Elimination~\citep{zhang2024agentprune,wang2025agentdropout}{,} \\
        \ \ Learned Topology Generation~\citep{zhang2024gdesigner,li2025argdesigner,jiang2025gtd}{,} \\
        \ \ Debate Efficiency and Selective Participation~\citep{zeng2025s2mad,wu2026debateocr}{,} \\
        \ \ System-Level Routing and Budget-Aware Coordination~\citep{yue2025masrouter,zhou2025mass,jin2025corl}{.}
        , leaf3, text width=45em, section_4, draw=tree-cyan]
    ]
    [\textbf{\ \ \ \ \ \ \ \ \ Agent Communication and} \\ \ \textbf{\ \ \ \ \ \ \ Interaction Optimization}~(\S\ref{subsec:mas_communication}), section_4, draw=box-cyan,  text width=18em
        [\ \ Message-Level Communication Compression~\citep{chen2025optima,yang2025codeagents}{,}
        Runtime Resource Allocation~\citep{gandhi2024budgetmlagent,supervisoragent2025}{.} 
        , leaf3, text width=45em, section_4, draw=tree-cyan]
    ]
    [\textbf{\ \ \ \ \ \ \ \ Computation Efficiency}~(\S\ref{subsec:mas_compute}), section_4, draw=box-cyan,  text width=18em
        [\ \ Cross-Context KV Cache Reuse~\citep{ye2025kvcomm,kriuk2025qkvcomm,bian2026tokendance}{,} \\
        \ \ Multi-Adapter and Cross-Model Cache Sharing~\citep{jeon2026lragent,liu2024droidspeak}{.}
        , leaf3, text width=45em, section_4, draw=tree-cyan]
    ]
    [\textbf{\ \ \ \ \ \ \ \ \ Memory Architecture and} \\ \textbf{\ \ \ \ \ \ \ \ \ Retrieval Efficiency}~(\S\ref{subsec:mas_memory}), section_4, draw=box-cyan,  text width=18em
        [Memory Topologies~\citep{wu2025memory,srmt2024,legomem2024,zou2025latentmas,yu2026multi}{,}
	Role-Specific and Self-Organizing Memory~\citep{yuen2025intrinsic,evocf2026}{,} \\
	\ \ Token-Budget-Aware Retrieval and Capacity Control~\citep{rcr2025,gmemory2025,agentnet2025}{.}
        , leaf3, text width=45em, section_4, draw=tree-cyan]
    ]
]
[\ \ \ \ \ \ \ \ Token Economics of \\ Intelligent Agent Ecosystems~(\S\ref{sec:ecosystem}), text width=10em, section_5, draw=box-green, text width=15em
    [\textbf{\ \ \ \ \ \ \ \ \ \ \ \ \ Problem Modeling:} \\ \textbf{\ \ \ Ecosystem Token Economics}~(\S\ref{subsec:eco_problem}), section_5, draw=box-green,  text width=18em
    	[\ \ Time-allocation Framing~\citep{becker1965allocation}{,} 
            Queueing-economics and Priority Pricing~\citep{afeche2004pricing,marchand1974priority,walters1961congestion}{,} \\
            \ \ Welfare Tradition~\citep{arrow1962economic,sen1970collective}{,} 
            Evolutionary-economics Sense~\citep{nelson1985evolutionary}{.},
            leaf3, text width=45em, section_5, draw=tree-green]
    ]
    [\textbf{\ \ \ \ \ \ \ \ \ Producer-Consumer Interaction:} \\ \textbf{\ \ \ \ \ \ \ \ Pricing and Congestion}~(\S\ref{subsec:eco_producer_consumer}), section_5, draw=box-green,  text width=18em
        [\ \ Scheduling and Prefill-decode Separation~\citep{yu2022orca,agrawal2024splitwise,zhong2024distserve,agrawal2024sarathi}{,} \\
        \ \ KV-cache Hierarchies and Prompt-caching Pricing~\citep{kwon2023vllm,zheng2024sglang,qin2025mooncake}{.}, leaf3, text width=45em, section_5, draw=tree-green]
    ]
    [\textbf{\ \ \ \ \ \ \ \ \ Producer-Producer Rivalry:} \\ \textbf{\ \ \ \ \ \ \ \ \ Oligopoly and Moats}~(\S\ref{subsec:eco_producer_producer}), section_5, draw=box-green,  text width=18em
        [\ \ Cost-reduction Mechanisms~\citep{frantar2023optq,awq2024,leviathan2023speculative}{,} 
        Open-weight Outside Options~\citep{deepseekv3,touvron2023llama}{,} \\
        \ \ Lock-in Vectors~\citep{anthropic2024mcp}{.}, leaf3, text width=45em, section_5, draw=tree-green]
    ]
    [\textbf{\ \ \ \ \ \ \ \ \ Regulator-Market Interaction:} \\ \textbf{\ \ \ \ \ \ \ Internalizing Externalities}~(\S\ref{subsec:eco_regulator_market}), section_5, draw=box-green,  text width=18em
        [\ \ Alignment Tax and Safety Pipelines~\citep{inan2023llamaguard,rebedea2023nemo,dong2024safeguard}{,} \\
        \ \ Privacy-preserving Inference Architectures~\citep{dong2023puma,li2022mpcformer}{,}
        Green Serving Efficiency~\citep{xu2025resource,leviathan2023speculative,you2023zeus}{.}, leaf3, text width=45em, section_5, draw=tree-green]
    ]
    [\textbf{\ \ \ \ \ \ \ \ \ Towards a Dynamic Token} \\ \textbf{\ \ \ \ \ \ \ \ Ecosystem Adjustment}~(\S\ref{subsec:eco_dynamic}), section_5, draw=box-green,  text width=18em
        [\ \ Jevons Dynamic in Token Ecosystems~\citep{jevons1865coal,agrawal2024splitwise,zhong2024distserve,qin2025mooncake,zheng2024sglang}{.}
        , leaf3, text width=45em, section_5, draw=tree-green]
    ]
]
[\ \ \ \ \ \ A Security Perspective on \\ \ \ \ \ \ \ \ \ Token Economics~(\S\ref{sec:eco_security}), appendix, draw=box-yellow, text width=15em
	[\textbf{\ \ \ \ \ \ \ \ \ \ Risk Categories Along the} \\ \textbf{\ \ \ \ \ \ \ \ \ \ \ \ \ Token Lifecycle}~(\S\ref{sec:security_taxonomy}), appendix, draw=box-yellow,  text width=18em
            [\ \ Input-Token Risk~\citep{zou2023universal,greshake2023indirect,anil2024manyshot}{,} 
            External-Token Risk~\citep{zou2025poisonedrag}{,}
            Internal-Token Risk~\citep{hubinger2024sleeper}{,} \\
            \ \ Inter-Agent Token Risk~\citep{debenedetti2024agentdojo,fang2024oneday}{,}
            Market-Level Token Risk~\citep{dong2025an}{.}
            , leaf2, text width=45em, appendix, draw=tree-yellow]
        ]
        [\textbf{\ \ \ \ \ \ \ \ Empirical Security Channels:} \\ \textbf{\ \ \ \ \ \ Evidence and Mechanisms}~(\S\ref{sec:security_empirical}), appendix, draw=box-yellow,  text width=18em
            [\ \ MVerification Costs Alter the Shadow Price of Tokens~\citep{liu2024prompt_injection}{,} \\
            \ \ Agentic Actions Transform Tokens into High-Stakes Outputs{,} \\
            \ \ Confidentiality Constraints Increase Communication Overhead{.}
            , leaf2, text width=45em, appendix, draw=tree-yellow]
        ]
        [\textbf{\ \ An Economic Cost Model Under Security Constraints}~(\S\ref{sec:security_model}), appendix, draw=box-yellow, text width=65em
        ]
        [\textbf{\ \ Policy Implications: Governance as Economic Infrastructure}~(\S\ref{sec:security_policy}), appendix, draw=box-yellow, text width=65em]
]
[\ \ Trends and Opportunities~(\S\ref{sec:future}), section_6, draw=box-purple, text width=15em
    [\textbf{\ \ \ \ \ \ \ \ \ \ \ \ \ \ \ \ \ \ Major Trends} \\ \textbf{\ \ \ \ \ \ \ \ \ \ \ in Token Economics}~(\S\ref{subsection:future-trends}), section_6, draw=box-purple,  text width=18em
        [\ \ Efficient Agent Inference and System Design~\citep{he2026harness}{,} \\
        \ \ Adaptive and Budget-Aware Token Allocation~\citep{yang2026dynamicearlyexit,jiang2025flashthink,jeong2024adaptive,gandhi2024budgetmlagent,jin2025corl}{,} \\
        \ \ Memory as Durable Capital with Compounding Returns~\citep{park2023generative,shinn2023reflexion,wang2023voyager,xu2025amem}{,} \\
        \ \ From Textual to Representational Token Exchange~\citep{hao2025coconut,amos2026thinkingstates,kriuk2025qkvcomm,wu2026debateocr}{,} \\
        \ \ Security Overhead as an Endogenous Efficiency Constraint{,} More Cost-Effective Hardware Chips{.}, leaf4, text width=45em, section_6, draw=tree-purple] 
    ]
    [\textbf{\ \ \ \ \ \ \ \ \ \ \ Emerging Opportunities} \\ \textbf{\ \ \ \ \ \ \ \ \ \ \ for Token Economics}~(\S\ref{subsection:future-opportunities}), section_6, draw=box-purple,  text width=18em
        [\ \ Differentiable Token Budgeting~\citep{yang2026dynamicearlyexit,jiang2025flashthink,jin2025corl}{,} \\
        \ \ Standardized Benchmarking and Cost Attribution~\citep{wang2025agenttaxo,salim2026tokenomics,zhu2025multiagentbench}{,} \\
        \ \ Real-Time Token Markets and Dynamic Pricing{,} 
        \ \ Token-Level Scaling Laws~\citep{kaplan2020scaling,hoffmann2022training,zhu2025multiagentbench,kim2025towards}{,} \\
        \ \ Security-Aware Token Budgeting{.}, leaf4, text width=45em, section_6, draw=tree-purple]
	]
]
]
	\end{forest}
}
\caption{Organizational taxonomy of the survey. This roadmap outlines our dual-view exploration of LLM agent token economics, categorizing consumption, efficiency, and economic models across single-agent, multi-agent, and ecosystem scales, while addressing security and future directions.}
\label{fig:all}
\end{figure}
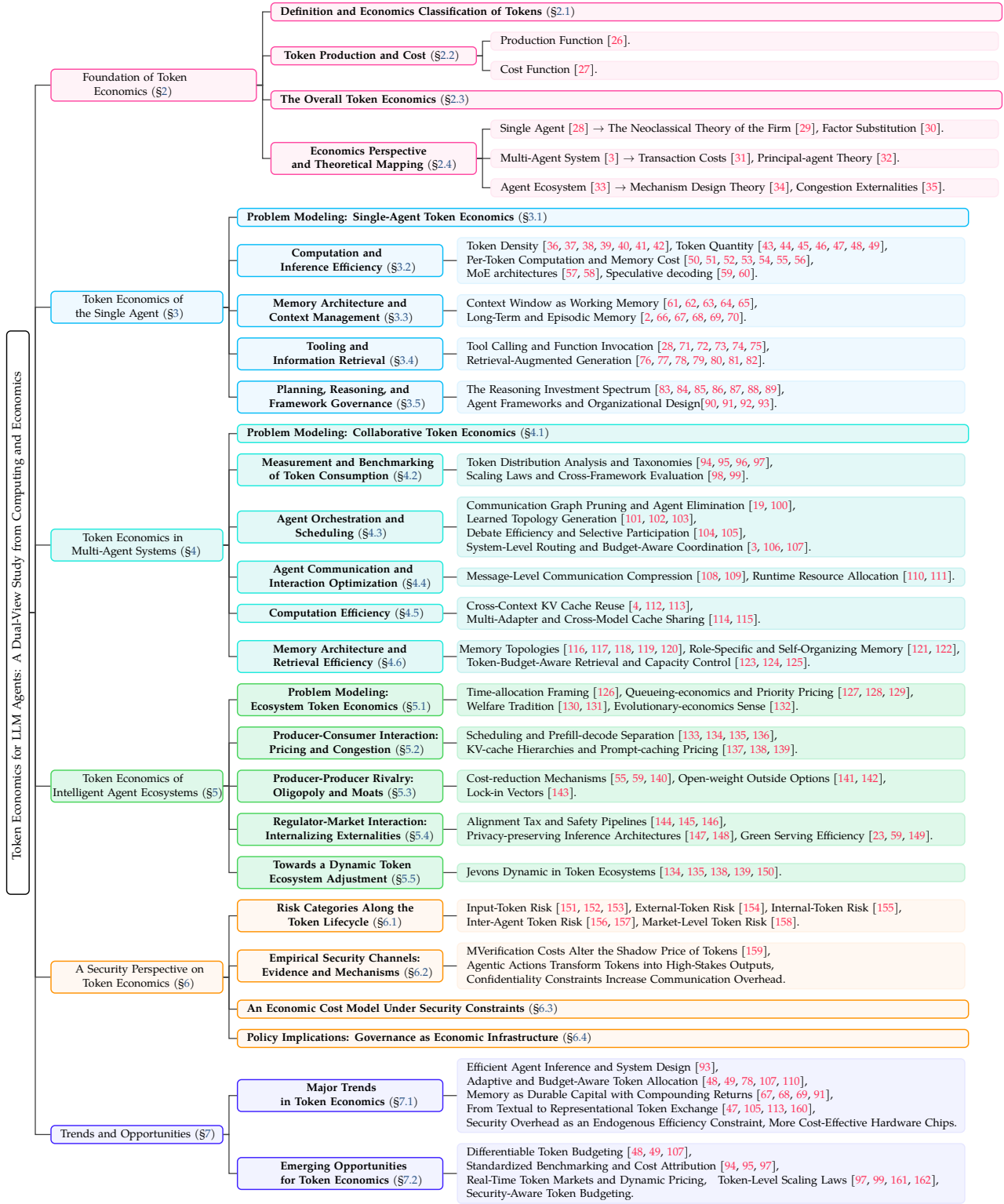

To systematically investigate the intersection of computer science and economics, the remainder of this paper follows a rigorous, progressive structure (see~\Cref{fig:all}). \Cref{sec:concepts} establishes the theoretical foundations of token economics, including the economic classification of tokens, token production and cost formulations, and the mapping from agent architectures to classical economic theories. Building on this foundation, \Cref{sec:single_agent} examines single-agent optimization through the lens of cost compression, factor substitution, and budget-aware reasoning under finite resource constraints. \Cref{sec:multi_agent} then elevates the discussion to collaborative LLM-agent systems, analyzing communication topology, coordination efficiency, and the mitigation of internal transaction friction. \Cref{sec:ecosystem} transitions to the macro-level agent ecosystem, where we review congestion scheduling, market clearing, and mechanism design for resource allocation in multi-tenant environments. \Cref{sec:eco_security} further introduces a security economics perspective, showing how threats and defenses reshape token utility and implicit system costs. Finally, \Cref{sec:future} outlines future directions spanning theoretical, systems, and societal dimensions, and \Cref{sec:conclusion} closes the paper.

\clearpage
\section{Foundations of Token Economics}
\label{sec:concepts}

This section establishes the theoretical foundation for token economics. \Cref{sec:concepts_classification} defines tokens’ triple economic attributes and their economic classification along the inference lifecycle. \Cref{sec:concepts_production} and \Cref{sec:concepts_te} formulate the token production function and cost structure, unifying engineering, resource allocation, and security under a dual-optimization framework. \Cref{sec:concepts_organizational} then maps single-agent, MAS, and ecosystem scales to organizational economics theories. Together, these three parts provide the economic lens used throughout the remainder of the survey.

\subsection{Definition and Economic Classification of Tokens}
\label{sec:concepts_classification}

In the traditional context of LLM agents, a token is narrowly defined as the fundamental unit of information processing. It serves as the minimal data structure that translates human semantics into computable representations~\cite{ahia2023languages}. Economically, however, the token has moved beyond this technical role as AI paradigms evolve toward complex commercial ecosystems. As we will see below, it now exhibits a \textit{triple economic attribute}.

\remarkbox{\label{rmk:triple_eco}
A \textbf{Triple Economic Attribute} of LLM Token:
\begin{itemize}[leftmargin=*]
    \item \emph{Factor of Production} (\textsf{Capital Goods~\cite{boldrin2024theory}}): Producing tokens demands tangible consumption of computational capital (GPUs, memory, power) and time. 
    \item \emph{Medium of Exchange} (\textsf{Currency~\cite{xu2024real}}): 
    Although dynamically generated on demand, the strict computational cost of their production and the ubiquitous practice of billing by token establish tokens as the de facto currency of AI liquidity~\footnotemark.
    \item \emph{Unit of Account~\cite{brzezinski2024understanding}}: Task complexity and AI productivity are objectively quantified by token expenditure variables.
\end{itemize}
}
\footnotetext{Unlike perfectly fungible fiat currencies, tokens exhibit significant individual heterogeneity in their cognitive utility and information density.}

This tri-faceted attribute serves as the foundational cornerstone for our dual-view analysis:
\begin{itemize}
    \item \textbf{Factor of Production (Micro-Level):} Generating and processing tokens directly consumes physical computing capital, such as GPU memory bandwidth and electricity. Consequently, tokens act as essential intermediate inputs within the AI production process. This attribute forms the theoretical basis for our subsequent analysis of factor substitution and organizational friction in ``Single-Agent'' (\Cref{sec:single_agent}) and ``Multi-Agent'' (\Cref{sec:multi_agent}) architectures.
    
    \item \textbf{Medium of Exchange and Unit of Account (Macro-Level):} Because API ecosystems universally rely on per-token billing, the token has become the standard currency driving the AI economy, while simultaneously providing an objective metric to quantify task complexity and systemic AI productivity. These intertwined attributes legitimize the introduction of congestion scheduling and scarce-capacity allocation mechanisms in our ``Ecosystem'' analysis (\Cref{sec:ecosystem}).
\end{itemize}

While tokens share these overarching economic traits, they also exhibit heterogeneous economic behaviors depending on their origin and systemic role. Anchored in their primary role as a factor of production, \Cref{tab:token_classification} categorizes these tokens and outlines their corresponding economic properties. The specific correspondence of these tokens within the agent architecture is detailed in \Cref{fig:pipeline}, which illustrates their dynamic lifecycle, specifically how they are generated, exchanged, and consumed throughout the actual inference process.

\begin{table}[!htbp]
\centering
\scriptsize
\caption{Economic classification of tokens in agent ecosystems.}
\label{tab:token_classification}
\resizebox{\columnwidth}{!}{%
\begin{tabular}{llll}
\toprule
\textbf{Category} & \textbf{Characteristics} & \textbf{Economic Meaning} \\
\midrule
\textbf{Input Token} & Encoding of user prompts & Intermediate Products \\
\textbf{Reasoning Token} & Chain-of-Thought sequences & Intermediate Products \\
\textbf{Communication Token}& Context shared and negotiated across multi-agent systems & Intermediate Products \\
\textbf{External Token} & Context acquired via RAG or API calls & Intermediate Products \\
\textbf{Output Token} & Final model-generated responses delivered to the user & Total Industrial Output Value \\
\bottomrule
\end{tabular}
}
\end{table}

\subsection{Token Production and Cost}
\label{sec:concepts_production}

Before introducing the specific formulations, we emphasize that these models act as the theoretical engine for our entire dual-view analysis. This unified framework scales systematically from single-agent inference (\Cref{sec:single_agent}) to MAS (\Cref{sec:multi_agent}), ecosystem governance (\Cref{sec:ecosystem}), and endogenous security constraints (\Cref{sec:eco_security}).

The output answer quality ($Y$) of an agent is not a simple linear extrapolation of a single technical metric. Instead, it is jointly produced by multiple interconnected factors: the foundation model's innate technological endowment ($A$), computational capital ($K$), intermediate token consumption ($M$), and human-AI collaborative labor ($L$).

To construct a computable framework, we first formalize how these heterogeneous inputs are transformed into the value of the final model response through a generalized production function: $Y = A \cdot f(K, L, M) \cdot e^{\epsilon}$. Here, $e^{\epsilon}$ captures stochastic shocks such as sampling temperature and non-deterministic hallucinations. Before instantiating this generalized form, we must distinguish the two economic relationships between computational capital ($K$) and intermediate tokens ($M$): \textbf{\emph{substitutability}} and \textbf{\emph{complementarity}}.

On one hand, algorithmic design allows these factors to act as {\emph{substitutes}} while maintaining a constant answer quality ($Y$). For instance, a resource-constrained agent can leverage extensive Chain-of-Thought reasoning tokens (high $M$) to compensate for a smaller foundation model (low $K$). Conversely, a massive frontier model (high $K$) can reach the same correct response via zero-shot inference while expending far fewer tokens (low $M$). On the other hand, under the physical constraints of LLM agent inference, $K$ and $M$ also exhibit strong {\emph{complementarity}}. The marginal productivity of one factor is amplified by, and tied to, the availability of the other. For instance, processing large token contexts demands proportionally large KV-cache capacity and memory bandwidth. 

Because token economics operates along a spectrum between perfect substitution and rigid complementarity, we need a function that can capture this elasticity. We therefore instantiate the generalized form as a modified nested Constant Elasticity of Substitution (CES) production function~\cite{behrman2024tutoring}

\begin{equation}
\label{eq:token_econimics}
    Y = A \cdot [\delta K^\rho + (1-\delta)M^\rho]^{\frac{\theta}{\rho}} \cdot L^\beta \cdot e^{\epsilon}
\end{equation}
where 
\begin{itemize}
    \item $A$ (Total Factor Productivity): Acts as a global multiplier on $Y$; a superior model architecture raises the absolute ceiling of $Y$ for any given input.
    \item $\rho$ and $\delta$ (Substitution \& Distribution Parameters): $\rho$~\footnote{In the CES framework, the elasticity of substitution is defined as $\sigma = \frac{1}{1-\rho}$. As $\rho \to 1$ (where $\sigma \to \infty$), the factors become perfect substitutes, implying that token accumulation can fully offset computational deficits. Conversely, as $\rho \to -\infty$ (where $\sigma \to 0$), the factors exhibit rigid complementarity. This extreme boundary characterizes the "Memory Wall" in LLM inference, where forcing additional token generation without sufficient hardware capacity (K) yields negligible cognitive output and triggers out-of-memory failures.} governs the elasticity of substitution, determining how seamlessly tokens can offset compute deficits to maintain a constant $Y$, while $\delta$ dictates the relative weight of physical compute versus token volume in producing $Y$.
    \item $\theta$ and $\beta$ (Returns to Scale): Dictate whether scaling machine inputs and human labor, respectively, yields increasing or diminishing marginal gains in $Y$.
\end{itemize}

Having formulated the token production function, we can conceptualize LLM agent inference not merely as a technical pipeline, but as a constrained optimization problem. The system seeks to maximize the output answer quality while adhering to a predefined resource budget. 
To evaluate economic viability, we construct the system's \textbf{Cost Function ($TC$)}, which quantifies the total economic expenditure of all factor inputs during a given inference lifecycle~\cite{raval2023testing}:
\begin{equation}
    TC = P_k \cdot K + P_m \cdot M + w \cdot L,
\end{equation}
where
\begin{itemize}[leftmargin=*]
    \item $P_k$ denotes the rental price of physical computational capital (e.g., GPU depreciation).
    \item $P_m$ represents the procurement price per intermediate token (e.g., API billing rates).
    \item $w$ denotes the wage rate of human cognitive labor, capturing the opportunity cost of human participation. It quantifies the implicit value of time and cognitive bandwidth expended by the user during human-in-the-loop interactions, such as prompt engineering and multi-turn alignment.
\end{itemize}

\subsection{The Overall Token Economics}
\label{sec:concepts_te}
Building upon the production and cost structures, we formalize the LLM agent inference process as a rigorous constrained resource-allocation problem. The system's objective is to minimize the total cost ($TC$) subject to an answer quality ($Z$). The overall Token Economics can be elegantly expressed as: 

\begin{equation}
\label{eq:te}
    \min TC  \quad s.t. \quad Y \ge Z.
\end{equation}

Having unified token economics into a single scalarized objective function (\Cref{eq:te}), we can systematically organize the LLM literature. We categorize existing research into three orthogonal paradigms, each targeting a different component of the theoretical model:

\begin{enumerate}[label=\textbf{Paradigm \Alph*}, align=left, leftmargin=*, itemsep=6pt]
    \item \textbf{\textsc{Engineering Optimization}} (Optimizing System Parameters): 
    This paradigm focuses on expanding the foundational production frontier. By introducing architectural innovations (e.g., MoE~\cite{fedus2022switch}) to elevate Total Factor Productivity ($A$) and systems-level optimizations (e.g., prompt caching) to compress unit prices ($P_k, P_m$), these works shift the physical limits of the system, raising the absolute ceiling of $Y$ while minimizing baseline costs.
    
    \item \textbf{\textsc{Resource Allocation}} (Optimizing Control Variables): 
    This paradigm focuses on dynamic execution under quality constraints ($Y \ge Z$). It explores how inference agents intelligently route and balance physical compute ($K$) and heterogeneous token inputs (e.g., \emph{uncached} versus \emph{cached}) to minimize total cost ($TC$) before hitting the threshold of diminishing marginal returns.

    \item \textbf{\textsc{Security Management}} (Bounding Stochastic Noise): 
    This paradigm focuses on mitigating severe negative externalities. Adversarial attacks introduce extreme volatility to the disturbance term ($e^{\epsilon}$), causing the output quality $Y$ to catastrophically collapse. This line of research treats defense as an endogenous economic constraint, aiming to bound expected utility loss without disproportionately inflating the inference cost $TC$. 
\end{enumerate}

\subsection{Economics Perspective and Theoretical Mapping}
\label{sec:concepts_organizational}

\begin{figure}[!htb]
  \centering
  \includegraphics[width=0.98\linewidth]{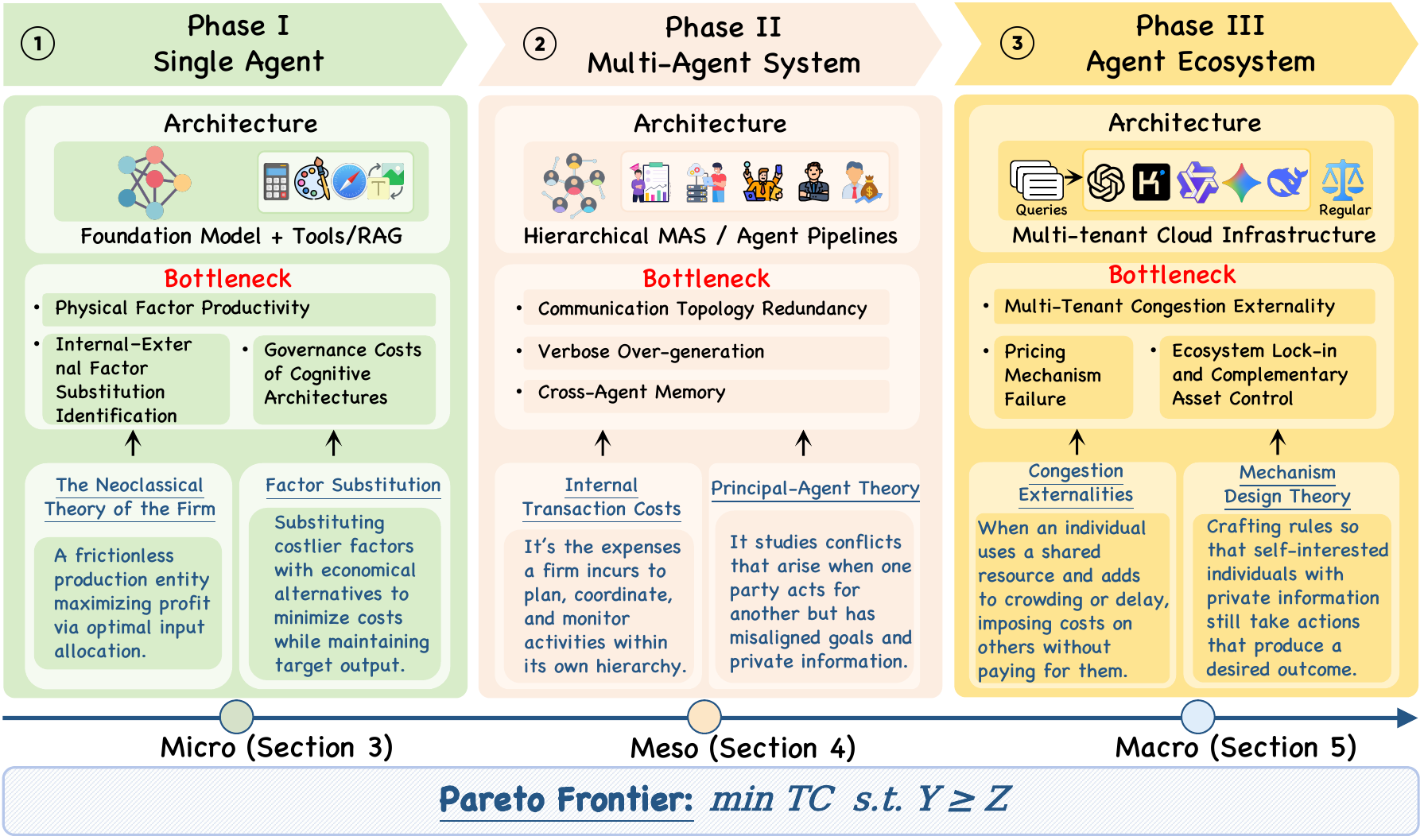}
  \caption{Isomorphic Mapping between Agent Architectures and Economics. The scaling of LLM agents from single nodes to open ecosystems strictly mirrors the economic evolution of a sole proprietorship, a hierarchical corporate organization, and a multi-sided platform market. The unified objective across all scales is achieving the Pareto frontier ($min\ TC\ s.t.\ Y\ge Z$) under quality constraints. }
  \label{fig:map_concepts}
\end{figure}

As LLM agent architectures scale to handle increasingly complex tasks~\cite{li2024survey}, existing optimization frameworks remain largely confined to engineering heuristics and physical metrics. This systems-centric perspective does not fully capture the economic realities of resource allocation.
We argue that the architectural progression of LLM-agent systems exhibits a strict structural isomorphism with the evolution of human economic systems: evolving from standalone firms, to corporate hierarchies, and ultimately to multi-sided platform markets~\cite{brynjolfsson2023information}.
To bridge this theoretical gap, we elevate token efficiency from heuristic hardware tuning to rigorous economic mechanism design. Through this dual-view lens, we structure the remainder of the survey around three progressive phases of organizational complexity (\Cref{fig:map_concepts}):
\begin{itemize}[leftmargin=*, itemsep=4pt, topsep=4pt]
    \item \textbf{Phase I: The Single Agent (Micro-Level):} Analogous to the neoclassical firm, this phase focuses on factor substitution and internal resource routing~\cite{schick2023toolformer,chodorow2025neoclassical,hubmer2026not}.
    \item \textbf{Phase II: Multi-Agent Systems (Meso-Level):} Mirroring corporate hierarchies, this phase addresses the internal transaction costs, communication frictions, and principal-agent alignment challenges inherent in distributed agent networks~\cite{yue2025masrouter,patil2024firm,lavi2022principal}.
   \item \textbf{Phase III: Agent Ecosystems (Macro-Level):} Functioning as multi-tenant platform markets, this phase explores mechanism design and dynamic pricing to mitigate congestion externalities and allocate scarce serving capacity under service and capacity constraints~\cite{basu2023stablefees,pycia2023theory,ershov2024variety}.
    
\end{itemize}

\subsubsection{Phase I: Single Agent Perspective}
\label{sec:phase1}
\sectionsubtitle{The Neoclassical Theory of the Firm~\cite{chodorow2025neoclassical} and Factor Substitution~\cite{hubmer2026not}}

As shown in~\Cref{fig:map_concepts}, the basic architecture of a single agent consists of a foundation model combined with external tools. This design closely parallels the setup of a sole proprietorship in neoclassical economics~\cite{varian1992micro}. Just as a firm achieves its objectives by improving production technology and optimizing factor allocation, an agent must not only enhance its internal productivity through technical upgrades, but also coordinate the use of external tools during operation. Its core objective is to minimize total computational cost, subject to a given output-quality constraint, that is, a specific level of cognitive output.

Under neoclassical theory, an agent’s productive capacity is governed by two core economic principles. First, to raise baseline capacity, the agent shifts its production possibility frontier outward by introducing technological strategies. Second, like the production function of a physical firm, the agent’s internal capability has a rigid production possibility frontier, constrained by knowledge truncation and logical bottlenecks. To overcome this limit, the agent decides between relying on internal reasoning and seeking external calls. In neoclassical terms, external tool use can be regarded as factor substitution. When internal generation faces very high marginal cost or declining accuracy, the agent obtains external tool tokens $M_{ext}$ to efficiently replace expensive or inefficient internal production factors.

Finally, all factor inputs, whether internally generated reasoning tokens $M_{int}$ or external tool-call tokens $M_{ext}$, are subject to resource scarcity. To reconcile the latency and formatting overheads associated with external tools, we introduce the economic concept of the shadow price $\tilde{P}$~\footnote{In microeconomic theory, a shadow price ($\tilde{P}$) represents the marginal value of relaxing a specific constraint. Its mathematical formulation evolves dynamically as the system organization scales, establishing a unified theoretical framework that spans from micro-scale inference to macro-scale ecosystems. Specifically, in \Cref{sec:single_agent}, the shadow price is defined as $\tilde{P}_{int/ext} = P_m + w \cdot \tau_{inf}$, where $P_m$ denotes the per-token procurement price, and $\tau_{inf}$ accounts for inference latency or tool-invocation overhead. The term $w \cdot \tau_{inf}$ internalizes the opportunity cost of time, transforming temporal latency into a tangible economic expenditure. This formulation is extended in \Cref{sec:multi_agent} to $\tilde{P}_{comm} = P_m + w \cdot \tau_{sync} + \Delta C_{coord}$, which incorporates inter-agent synchronization latency ($\tau_{sync}$) and the coordination costs associated with format alignment ($\Delta C_{coord}$). Finally, in \Cref{sec:ecosystem}, the model converges to $\tilde{P}_{eco} = P_m + w \cdot \tau_{cong} + C_{comp}$, further encompassing congestion latency ($\tau_{cong}$) induced by multi-tenant competition, as well as costs related to system compliance and environmental externalities ($C_{comp}$).}. The agent’s operation is therefore abstracted as a classic cost-minimization dual problem. In a dynamic environment, it must precisely balance the marginal rate of technical substitution according to the shadow prices of the different factors. By jointly evaluating input-output efficiency, the agent can approach the Pareto frontier under a fixed quality constraint. We present the problem modeling of the single-agent token economics and relevant techniques in~\Cref{sec:single_agent}.


\subsubsection{Phase II: Multi-Agent System Perspective}
\label{sec:phase2}
\sectionsubtitle{Transaction Costs~\cite{patil2024firm} and Principal-agent Theory~\cite{lavi2022principal}}

As LLM architectures evolve from a single agent to multi-agent systems (MAS), the system shifts from a sole proprietorship to a hierarchical corporate organization. Just as firms improve productivity through division of labor and coordination, MAS relies on agent specialization. Its core objective is to minimize total cost under a target level of collective cognitive output.

In organizational economics, the capabilities of MAS are mainly governed by two principles. First, specialization gains are maximized by expanding the organizational boundary through specialized roles and topology. Second, like the expansion of a physical firm, MAS also faces rigid scaling bottlenecks constrained by internal transaction costs and the Principal-Agent dilemma~\cite{jensen1976theory,lavi2022principal}. Following Williamson’s framework~\cite{williamson1985economic,patil2024firm}, we formalize the overhead of inter-agent state synchronization, repeated context transmission, and rigid JSON alignment not as mere execution loss, but as internal transaction costs. In economic terms, these ``communication taxes'' represent the friction required to maintain coherence across fragmented organizational units~\cite{patil2024firm,wang2025agenttaxo}.

Ultimately, all tokens, whether productive node tokens or transactional synchronization tokens, can be modeled as structural overhead. This allows MAS orchestration to be formulated as a classic Coasian boundary optimization and cost minimization problem. In a dynamic environment, the system must balance the marginal gains from specialization against the marginal transaction costs required to maintain consistency. Under a fixed quality constraint, MAS can then approach the Pareto efficient frontier of collaboration. We present the problem modeling of the MAS token economics and relevant techniques in~\Cref{sec:multi_agent}.


\subsubsection{Phase III: Agent Ecosystem Perspective}
\label{sec:phase3}
\sectionsubtitle{Mechanism Design~\cite{pycia2023theory} Theory and Congestion Externalities~\cite{ershov2024variety}}

As LLM architectures expand to macro-level ecosystems, the paradigm transitions from a hierarchical corporation to a multi-sided platform market. Just as open markets coordinate decentralized supply and demand, an agent ecosystem must orchestrate multi-tenant competition for shared cloud infrastructure. Its core objective is to allocate scarce serving capacity across heterogeneous users and providers, minimizing generalized ecosystem cost under service-level and capacity constraints while mitigating systemic resource contention.

Ecosystem efficiency is governed by two principles. First, it maximizes serving capacity by expanding the physical production frontier through supply-side infrastructure innovations like continuous batching. Second, this shared capacity confronts severe macro-bottlenecks: congestion externalities~\cite{walters1961congestion,basu2023stablefees} and market failures. When uncoordinated tenants fiercely compete for finite GPU memory, individual token hoarding inflicts queuing delays on others—manifesting as a computational tragedy of the commons.

Ultimately, token-mediated computational capacity is a strictly scarce resource. By formalizing these systemic frictions as market failures, ecosystem governance becomes a mechanism-design and dynamic-pricing problem~\cite{pycia2023theory}. To approach a sustainable frontier, platforms and regulators must deploy incentive-compatible interventions that induce agents to internalize congestion costs, thereby correcting externalities that decentralized heuristics cannot capture. We present the problem formulation of token economics in agent ecosystems and the corresponding techniques in~\Cref{sec:ecosystem}.

\clearpage
\section{Token Economics of the Single Agent}
\label{sec:single_agent}

\chapterepigraph
  {The first principle of Economics is that every agent is actuated only by self-interest.}
  {Francis Ysidro Edgeworth}
  {Mathematical Psychics, London: C. Kegan Paul, 1881, p.~16.}

To systematically deconstruct the token economics of single-agent architectures, this section proceeds from theoretical modeling to engineering optimization. 
First, \Cref{subsec:single_problem} formalizes the budget-constrained optimization problem for single-agent token economics, framing the dynamic substitution between internal reasoning and external tool tokens. 
Next, \Cref{subsec:single_compute} examines inference efficiency, demonstrating how to expand the production frontier by maximizing token information density and compressing unit computational costs. 
Finally, \Cref{subsec:single_memory}, \Cref{subsec:single_tool}, and \Cref{subsec:single_planning} analyze resource allocation across tool invocation, memory, and cognitive planning, detailing the algorithmic strategies necessary to approach the Product-Cost Pareto frontier.

\subsection{Problem Modeling: Single-Agent Token Economics}
\label{subsec:single_problem}

\leadparagraph{Definition.}
Building upon the neoclassical model established in~\Cref{sec:phase1}, single-agent token economics is fundamentally a constrained resource allocation problem. During execution, an agent must balance two computational factors with heterogeneous cost structures: (1) Internal reasoning tokens ($M_{int}$), which are generated via parametric memory and incur the shadow price of $\tilde{P}_{int}$. (2) External tool tokens ($M_{ext}$), which incur the shadow price of $\tilde{P}_{ext}$.

The total economic expenditure for single-agent inference is explicitly defined as $TC = P_k \cdot K + (\tilde{P}_{int} \cdot M_{int} + \tilde{P}_{ext} \cdot M_{ext}) + w \cdot L$. To achieve economic rationality, the Pareto frontier can be formalized as:
\begin{equation}
    \min_{K, L, M_{int}, M_{ext}} TC \quad \text{s.t.} \quad Y \ge Z.
\end{equation}

\begin{figure}[!htbp]
  \centering
  \includegraphics[width=0.9\linewidth, height=0.35\textheight, keepaspectratio]{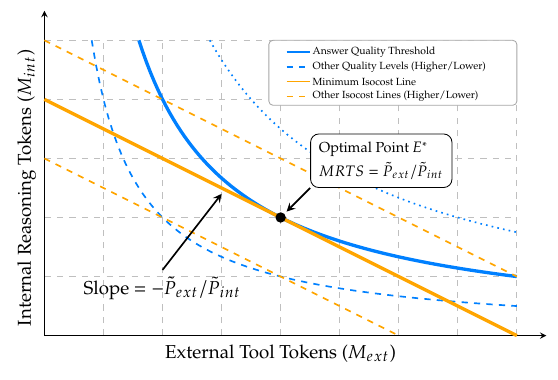}
  \caption{Single-agent resource routing as a constrained optimization problem. The optimal policy occurs at $E^*$, where the agent perfectly balances internal parametric reasoning ($M_{int}$) and external tool use ($M_{ext}$) to minimize total cost ($TC$) while strictly satisfying the minimum output quality constraint ($Y \ge z$).}
  \label{fig:single_agent}
\end{figure}

\leadparagraph{Example.}
To intuitively illustrate this routing logic, consider an agent tasked with real-time financial market analysis~\Cref{fig:single_agent}. To achieve a target output quality (represented by the single isoquant $Y$), the system must balance internal reasoning ($M_{int}$) and external retrieval ($M_{ext}$). If the agent heavily biases towards internal reasoning to reach $Y$, compensating for outdated or hallucinated facts requires disproportionate computational effort, which pushes the system onto a higher, suboptimal total cost ($TC$) curve. Conversely, an over-reliance on external retrieval leads to massive API payloads that inflate context processing and latency overheads, similarly escalating the total cost. The optimal operating point ($E^*$) is achieved where the $Y$ contour is tangent to the lowest possible isocost line ($TC^*$). At this cost-minimizing equilibrium, the agent strategically queries a single, high-density API for real-time stock prices and synthesizes the data internally, deliberately avoiding superfluous API calls that would merely inflate costs without being strictly necessary to satisfy the target quality threshold.

To solve this complex allocation problem, contemporary research advances along two macro-paradigms (see \Cref{sec:concepts_te}):
\begin{itemize}[leftmargin=*, itemsep=4pt, topsep=4pt]
    \item \textbf{Paradigm A: \textsc{Engineering Optimization}}. This approach physically alters the diagram by compressing baseline compute prices ($P_k, \tilde{P}_{int}$) (\Cref{subsec:single_compute}) or amortizing external retrieval costs into reusable memory capital (\Cref{subsec:single_memory}), effectively pushing the budget line outward.
    \item \textbf{Paradigm B: \textsc{Resource Optimization}}. This approach focuses on algorithm design. By reducing tool integration friction (\Cref{subsec:single_tool}) and deploying marginal scheduling (\Cref{subsec:single_planning}), it actively routes the agent's execution path directly toward $E^*$.
\end{itemize}

\Cref{tab:single_eco_mapping} summarizes how subsequent algorithmic designs leverage these two paradigms to approach the Pareto frontier of single-agent token economics.

\begin{sidewaystable}
\caption{Single-agent: technical solutions and economic mapping. (Paradigm A: \textsc{Engineering Optimization}; B: \textsc{Resource Allocation})}
\label{tab:single_eco_mapping}
\centering

\renewcommand{\arraystretch}{1.2} 

\resizebox{\linewidth}{!}{
\begin{tabular}{@{} 
    >{\raggedright\arraybackslash}p{3.8cm} 
    >{\raggedright\arraybackslash}p{4.2cm} 
    >{\raggedright\arraybackslash}p{9.4cm} 
    >{\raggedright\arraybackslash}p{9.4cm} 
    c
    c
    @{}}
\toprule
\multirow{2}{*}{\textbf{Chapter}} & \multirow{2}{*}{\textbf{Core Objective}} & \multirow{2}{*}{\textbf{Representative Technical Solutions}} & \multirow{2}{*}{\textbf{Economic Mapping}} & \multicolumn{2}{c}{\textbf{Paradigm (\Cref{sec:concepts_te})}} \\
\cmidrule(l){5-6}
& & & & \textbf{A} & \textbf{B} \\
\midrule
\multirow{3}{=}{Computation and Inference Efficiency \newline (\Cref{subsec:single_compute})} & 
\multirow{3}{=}{Cost Reduction and Efficiency Improvement: Reduce the generation cost per token and increase the information density of each token.} & 
Efficient Token
Embeddings: Dense Continuous Embeddings~\cite{bengio2003neural,mikolov2013efficient,pennington2014glove}, Subword Tokenization~\cite{sennrich2016neural,kudo2018sentencepiece}, Discrete Latent Tokenization~\cite{oord2017neural,razavi2019generating} & 
Elevate Total Factor Productivity (TFP, $A$) by increasing semantic density per factor input. & \checkmark & \xmark \\
\cmidrule(l){3-6}

& & 
Reducing the Number of Tokens: Chain-of-Thought~\cite{cheng2024compressed,shen2025codi,wang2025r1compress,xia2025tokenskip}, Latent Reasoning or Implicit Reasoning~\cite{hao2025coconut}, Early Exit~\cite{yang2026dynamicearlyexit,jiang2025flashthink} & 
Structural compression of internal factor ($M_{int}$) and dynamic truncation based on marginal cost-benefit analysis. & \checkmark & \checkmark \\
\cmidrule(l){3-6}

& & 
Lowering the Computational and Memory: Attention Optimization~\cite{dao2022flashattention,zaheer2020big,katharopoulos2020transformers}, KV Cache Optimization~\cite{li2024snapkv,zhang2023h2o}, 
Quantization~\cite{frantar2023optq}, Pruning~\cite{sanh2020movement}& 
Compress the effective capital cost ($P_k$) and internal token shadow price ($\tilde{P}_{int}$) by reducing compute, memory traffic, and execution latency. & \checkmark & \checkmark \\
\cmidrule(l){3-6}

& & 
Architectural Levers: MoE~\cite{fedus2022switch,zoph2022stmoe}, Speculative decoding~\cite{leviathan2023speculative,zhang2024draft} & 
Elevate TFP ($A$) via specialization and compress unit capital costs by reducing effective decoding latency. & \checkmark & \xmark \\
\midrule

\multirow{3}{=}{Memory Architecture and Context Management \newline (\Cref{subsec:single_memory})} & 
\multirow{3}{=}{Asset Accumulation: transform one-time context consumption into a reusable long-term system knowledge base.} & 
Working Memory: Lossless or Lossy Prompt Compression~\cite{jiang2023llmlingua,jiang2024longllmlingua,li2023selective}, Text Condensation and Extraction ~\cite{mu2023learning,chevalier2023autocompressors} & Amortize repetitive token expenditure into a one-time fixed investment, reducing variable material input ($M_{int}$). & \checkmark & \xmark \\
\cmidrule(l){3-6}

& & 
Storage Scheduling: Virtual Paging and Scheduling Between Internal and External Memory~\cite{packer2024memgpt} & 
Reduce the shadow price ($\tilde{P}_{ext}$) via dynamic paging and optimal state routing. & \checkmark & \checkmark \\
\cmidrule(l){3-6}

& & 
Episodic Memory: Error Reflection and Higher-order Cognitive Extraction~\cite{park2023generative,shinn2023reflexion}, Active Memory Pruning and Forgetting~\cite{zhong2024memorybank} & 
Accumulate reusable experiential knowledge to dynamically optimize future resource allocation and increase output quality ($Y$). & \xmark & \checkmark \\
\cmidrule(l){3-6}

& & 
Persistent and Structured Memory: Self-organizing memory graphs~\cite{xu2025amem}, Persistent memory-centric architectures~\cite{chhikara2025mem0} & 
Substitute variable token consumption (context-stuffing) with durable knowledge capital that yields compounding returns. & \checkmark & \checkmark \\
\midrule

\multirow{3}{=}{Tooling and Information Retrieval \newline (\Cref{subsec:single_tool})} & 
\multirow{3}{=}{Factor Substitution: under a limited budget, substitute between internal reasoning and external tools to minimize cost for a required output quality.} & 
Lowering Tool Integration Cost: MCP~\cite{hou2025model}, Tool Selection and Invocation Optimization~\cite{schick2023toolformer,patil2023gorilla,qin2024toolllm,du2024anytool,toolrl2024} & 
Compress integration friction ($\tilde{P}_{ext}$) and dynamically route tasks toward external tool use ($M_{ext}$). & \checkmark & \checkmark \\
\cmidrule(l){3-6}

& & 
Dynamic and Verified Retrieval: On-Demand Retrieval~\cite{selfrag2024,jeong2024adaptive}, Quality Verification of Retrieval and Tool Invocation~\cite{yan2024corrective}, Adaptive Retrieval Granularity~\cite{du2026arag}. & 
Approach the Pareto frontier by dynamically substituting between internal parametric reasoning ($M_{int}$) and external retrieval factors ($M_{ext}$). & \xmark & \checkmark \\
\cmidrule(l){3-6}

& & 
Structural Knowledge Acquisition: Structured Retrieval~\cite{edge2024graphrag,sarthi2024raptor,gao2023hyde} & 
Amortize indexing costs to increase the information density and capital leverage of external factors. & \checkmark & \xmark\\
\midrule

\multirow{3}{=}{Planning, Reasoning, and Framework Governance \newline (\Cref{subsec:single_planning})} & \multirow{3}{=}{Global Scheduling: plan exploration and trial-and-error paths in complex tasks, control total cost, and avoid getting stuck in loops.} & Reasoning Topologies Evolution: Linear Reasoning~\cite{cot2022,yao2023react}, Tree or Graph-structured Exploration Expansion~\cite{yao2023tree,besta2024graph}, Combining Monte Carlo Evaluation with Tree Search~\cite{zhou2024lats}, Explicit Plan Decomposition~\cite{wang2023plan}, Budget-Aware Search~\cite{li2026bavt} & Navigate the reasoning investment spectrum to preserve positive but diminishing returns to additional token spending. & \xmark & \checkmark\\
\cmidrule(l){3-6}

& & 
Framework and Skill Governance: Unified Agent Architectures~\cite{sumers2024cognitive}, Reuse of Code/Action Libraries~\cite{wang2023voyager}, Harness Layer Engineering~\cite{he2026harness} & 
Capitalize procedural knowledge to achieve economies of scale and govern long-run token allocation. & \checkmark & \checkmark \\
\cmidrule(l){3-6}

& & 
System Constraints: Error-loop Interruption and Terminal Optimization~\cite{yang2024sweagent}, Enforced Loop Budget Control & 
Enforce hard budget constraints (stop-loss) to intercept execution paths with negative marginal utility. & \xmark & \checkmark \\

\bottomrule
\end{tabular}
}
\end{sidewaystable}

\subsection{Computation and Inference Efficiency}
\label{subsec:single_compute}

An agent can improve token efficiency across the full token lifecycle, including token representation, generation, and consumption. 
This can be categorized along three progressively deeper dimensions: token density, token quantity, and the computation and memory cost associated with each token.


\paragraph{Token Density.} Token density refers to how each token can encode more useful information in a dense and learnable form.
Dense continuous embeddings (Word2Vec, GloVe~\cite{bengio2003neural,mikolov2013efficient,pennington2014glove}) replace sparse symbolic representations with compact semantic vectors that capture semantic and syntactic structure.
Subword tokenization (BPE, SentencePiece~\cite{sennrich2016neural,kudo2018sentencepiece}) balances vocabulary size, compositionality, and rare-word coverage through reusable subword units.
Discrete latent tokenization (VQ-VAE, VQ-VAE-2~\cite{oord2017neural,razavi2019generating}) compresses continuous multimodal inputs into compact discrete codes, preserving high information content under tight sequence budgets.
Together, these approaches raise total factor productivity $A$: each token carries more information, shifting the production frontier upward (\Cref{fig:single_agent}).


\paragraph{Token Quantity.} This dimension refers to how many tokens a model generates to complete a certain task.
\emph{Chain-of-Thought compression}~\cite{cheng2024compressed,shen2025codi,wang2025r1compress,xia2025tokenskip} shortens verbose reasoning traces into compact forms, directly reducing $M_{int}$.
\emph{Latent Reasoning}~\cite{hao2025coconut} moves reasoning from natural-language tokens into a compact continuous latent space.
\emph{Early Exit}~\cite{yang2026dynamicearlyexit,jiang2025flashthink} terminates generation once the marginal product of the next reasoning token falls below its marginal cost.


\paragraph{Per-token Computation and Memory.} This dimension decides the computational and memory cost behind each generated token.
\emph{Attention optimization}: FlashAttention~\cite{dao2022flashattention} eliminates IO-bound memory movement via tiled exact computation; BigBird~\cite{zaheer2020big} replaces dense attention with structured sparse patterns; linear attention~\cite{katharopoulos2020transformers} achieves linear complexity in sequence length.
\emph{KV cache optimization}: SnapKV~\cite{li2024snapkv} retains the most informative KV entries; H$_2$O~\cite{zhang2023h2o} keeps heavy-hitter tokens, which reduce the memory holding cost of decoding.
\emph{Quantization and pruning}: OPTQ~\cite{frantar2023optq} lowers weight precision while preserving quality; Movement Pruning~\cite{sanh2020movement} removes less useful weights, which reduces per-token compute and memory traffic.


Beyond density, quantity, and unit cost, there are other architectural levers that reduce per-token expenditure.
\emph{MoE architectures}~\cite{fedus2022switch,zoph2022stmoe} route each token to only a small expert subset, realizing the productivity gains of specialization while keeping per-token computation low.
\emph{Speculative decoding}~\cite{leviathan2023speculative,zhang2024draft} uses a small draft model to propose candidate tokens verified in parallel by the target model, reducing effective decoding cost.

\remarkbox{\label{rmk:single_compute}The three levers above act on distinct factors of the per-token production cost. \emph{Token density} raises factor productivity $A$---each token carries more useful information. \emph{Token quantity} reduces $M$ on the intensive margin by truncating reasoning traces and exiting early when the marginal product falls below the marginal cost. \emph{Per-token cost} attacks the unit price by reducing attention, KV-cache, and weight-precision overhead. MoE and speculative decoding combine specialization and parallelism to lower effective cost without sacrificing output quality.
}

\subsection{Memory Architecture and Context Management}
\label{subsec:single_memory}

An agent's memory system transforms token expenditure from single-use consumption into an \emph{investment--amortization} pattern, mapping onto the economic distinction between intermediate goods and \emph{capital goods}~\cite{varian1992micro}.

\leadparagraph{Context Window as Working Memory.}
The context window is a \emph{rival, excludable resource} contested by system prompts, tool schemas, conversation history, retrieved documents, and reasoning scratchpads. Admitting one additional retrieved token necessarily evicts one token of history---a \emph{constrained resource allocation} problem~\cite{varian1992micro} in which a fixed context budget must be partitioned across competing uses to maximize output.
Prompt-compression methods reduce the token cost of injected context: LLMLingua~\cite{jiang2023llmlingua} scores token importance with a small auxiliary model and drops low-utility tokens under a budget controller, while LongLLMLingua~\cite{jiang2024longllmlingua} additionally re-densifies and re-positions question-relevant content to mitigate the ``lost-in-the-middle'' failure mode.
Gisting~\cite{mu2023learning} and AutoCompressors~\cite{chevalier2023autocompressors} go further by training the model to fold prompts (or successive document segments) into cacheable soft tokens, converting recurring injection costs into a one-time investment.
Selective Context~\cite{li2023selective} takes a training-free route, pruning lexical units by self-information so that predictable (low-surprisal) tokens are discarded as redundant.
The optimality condition for context allocation can therefore be stated in words: token slots should be shifted across competing components until their marginal contribution to output is balanced. Liu et al.~\cite{liu2024lost} further show that this marginal product is \emph{position-dependent} (U-shaped attention)---analogous to spatial heterogeneity in land economics.

\leadparagraph{Long-Term and Episodic Memory.}
Beyond the working window, agent memory spans three tiers: \emph{working memory} (liquid capital, high turnover), \emph{long-term memory} (fixed capital, with write/retrieval/maintenance costs), and \emph{episodic memory} (intangible capital accumulated through \emph{learning-by-doing}~\cite{arrow1962economic}).
MemGPT~\cite{packer2024memgpt} implements OS-style virtual memory by treating the context window as fast main memory and external storage as slow disk, with explicit function-call \emph{interrupts} paging data between tiers.
Generative Agents~\cite{park2023generative} log experiences in a memory stream scored by recency, importance, and relevance, and periodically reflect to synthesize higher-level abstractions---an R\&D investment that lowers future retrieval costs along Arrow's learning curve.
Reflexion~\cite{shinn2023reflexion} performs verbal reinforcement without weight updates: after each failure, the agent stores a natural-language self-critique in an episodic buffer, priming subsequent trials with targeted error analysis.
MemoryBank~\cite{zhong2024memorybank} models retention via the Ebbinghaus Forgetting Curve, so that stale, low-significance entries are automatically retired (\emph{capital depreciation}) to prevent retrieval pollution.
A-MEM~\cite{xu2025amem} maintains a self-organizing memory graph in which each new note is linked to and updates semantically related entries, so the store grows denser rather than merely larger.
Mem0~\cite{chhikara2025mem0} pushes this design into production with a memory-centric architecture that continuously extracts, consolidates, and retrieves salient facts across sessions, providing persistent memory as a substitute for context-stuffing on the cost--quality frontier.

\remarkbox{\label{rmk:single_memory}Memory optimizations act on two levers. \emph{Friction cost compression}: paging (MemGPT), prompt compression (LLMLingua), and memory retirement reduce context-switching, injection, and retrieval-noise overhead. \emph{Factor productivity enhancement}: Reflection and error-to-skill conversion (Generative Agents, Reflexion) produce an Arrow-style learning curve in which accumulated experience lowers future average cost, yielding increasing returns that static production functions fail to capture.}

\subsection{Tooling and Information Retrieval}
\label{subsec:single_tool}

When an agent encounters a subtask, it chooses between internal parametric reasoning ($M_{int}$ at shadow price $\tilde{P}_{int}$) and external capability invocation ($M_{ext}$ at shadow price $\tilde{P}_{ext}$).
Both tool calling and RAG are instances of \emph{factor substitution}~\cite{varian1992micro,arrow1961capital}: the agent reallocates its input mix along the isoquant to minimize cost for a given output level.

\leadparagraph{Tool Calling and Function Invocation.}
Internal generation avoids integration overhead but is bounded by knowledge cutoffs; external invocation provides up-to-date results but incurs schema injection, call-parsing, and invocation-failure costs.
Toolformer~\cite{schick2023toolformer} learns optimal invocation timing via self-supervised training, letting a smaller model achieve large-model behaviour through tool \emph{capital leverage}.
Gorilla~\cite{patil2023gorilla} fine-tunes with retrieval-augmented API calling to suppress hallucination-induced retry costs.
ToolLLM~\cite{qin2024toolllm} scales to a large API repository through a neural retriever that controls schema-injection cost growth, while AnyTool~\cite{du2024anytool} introduces hierarchical retrieval (category $\to$ subcategory $\to$ API) that further reduces selection and injection overhead.
ToolRL~\cite{toolrl2024} replaces supervised fine-tuning with reinforcement-learning rewards over tool-use trajectories, internalizing \emph{when} and \emph{how} to invoke external APIs and thereby suppressing redundant calls that waste $M_{ext}$.
At the interface layer, unified protocols---OpenAI Function Calling and MCP~\cite{hou2025model}---standardize the tool--agent contract, lowering per-invocation integration cost and enabling plug-and-play tool composition.

\leadparagraph{Retrieval-Augmented Generation.}
At the agent level, RAG becomes an active \emph{factor allocation decision}: the agent dynamically decides \emph{whether}, \emph{when}, and \emph{how much} to retrieve.
Self-RAG~\cite{selfrag2024} uses reflection tokens to make retrieval conditional on need, while CRAG~\cite{yan2024corrective} adds a lightweight retrieval evaluator that triggers supplementary web search when confidence is low.
Adaptive-RAG~\cite{jeong2024adaptive} matches strategy to query complexity: simple queries skip retrieval entirely, complex ones invoke iterative multi-step retrieval.
On the indexing side, GraphRAG~\cite{edge2024graphrag} constructs entity knowledge graphs for global sensemaking and RAPTOR~\cite{sarthi2024raptor} builds recursive multi-level summaries; both amortize a high upfront indexing cost over many downstream queries.
HyDE~\cite{gao2023hyde} invests a small budget of ``hypothetical-document'' tokens per query to lift retrieval precision, trading a modest internal cost for sharper external matches.
A-RAG~\cite{du2026arag} exposes hierarchical retrieval primitives---keyword search, semantic search, and chunk read---directly to the agent, so retrieval granularity is chosen at inference time rather than fixed by the pipeline.

\remarkbox{\label{rmk:single_tool}The production cost of $M_{ext}$ decomposes into three token-denominated overheads: (i)~\emph{selection costs} (identifying the right tool or document), reduced by neural retrievers and hierarchical filtering; (ii)~\emph{formatting costs} (constructing the invocation), reduced by MCP and Function Calling; (iii)~\emph{verification costs} (confirming result validity), mitigated by CRAG. AnyTool and MCP compress $\tilde{P}_{ext}$ by targeting (i) and (ii) directly, shifting the MRTS equilibrium toward greater external token utilization. GraphRAG provides a canonical capital-leverage calculation: fixed indexing investment $I_{graph}$ is justified when $I_{graph}/Q < \Delta Y$ per query.
}

\subsection{Planning, Reasoning, and Framework Governance}
\label{subsec:single_planning}

The reasoning loop is the most token-intensive component of agent operation; strategy choice directly determines the \emph{intensity} of intermediate input investment.

\leadparagraph{The Reasoning Investment Spectrum.}
\label{subsubsec:reasoning}
Reasoning strategies span a spectrum of token investment, from minimal (direct prompting, CoT~\cite{cot2022}) through moderate (ReAct~\cite{yao2023react}, ToT~\cite{yao2023tree}, GoT~\cite{besta2024graph}) to heavy (LATS~\cite{zhou2024lats}). Each successive strategy raises total token expenditure while improving output quality, but with \emph{diminishing returns}: additional reasoning tokens remain useful, yet their incremental contribution eventually declines.
ReAct~\cite{yao2023react} \emph{interleaves} reasoning and acting so that each Thought is grounded by an Action and its Observation, suppressing hallucination at the cost of additional external tokens.
ToT~\cite{yao2023tree} generalizes CoT to a tree, generating and self-evaluating multiple candidate thoughts and searching with backtracking---abandoned branches represent sunk exploration costs traded for systematic coverage of the solution space.
GoT~\cite{besta2024graph} extends this to a directed graph in which thoughts can be \emph{merged}, \emph{distilled}, and \emph{refined}, enabling cross-branch synthesis unavailable in tree search.
LATS~\cite{zhou2024lats} embeds MCTS inside the agent loop, using an LM value function to score candidate nodes and self-reflections on failed paths as context for future iterations.
BAVT~\cite{li2026bavt} makes the search itself \emph{budget-aware}: a node-selection rule conditioned on the remaining-resource ratio interpolates smoothly between broad exploration and greedy exploitation as the token budget depletes.
Plan-and-Solve~\cite{wang2023plan} addresses missing-step errors in zero-shot CoT by first prompting an explicit numbered plan and then executing each subtask sequentially.

\leadparagraph{Agent Frameworks and Organizational Design.}
CoALA~\cite{sumers2024cognitive} draws on cognitive science (ACT-R, SOAR) to organize agents around modular memory stores, a structured action space (internal storage/retrieval; external execution/communication), and a generalized decision cycle---a unified lens for reasoning about token allocation across framework designs.
Voyager~\cite{wang2023voyager} couples an automatic exploration curriculum with an ever-growing skill library of executable code, illustrating skill-library economics: initial acquisition is expensive, but reuse cost is near-zero.
SWE-agent~\cite{yang2024sweagent} introduces a custom Agent--Computer Interface (ACI) with purpose-built commands for file navigation, search, and inline editing, showing that interface design can be as impactful as model capability in reducing per-action overhead.
Recent work formalizes this observation as the \emph{harness layer}~\cite{he2026harness}: the runtime scaffolding that mediates control flow, agency boundaries, and tool I/O is itself an economic design surface, and harness-level choices (loop budgets, retry policies, observation truncation) often dominate model-level differences in end-to-end token cost.
Hard loop budgets in particular act as stop-loss constraints: once the marginal return of an additional round falls below its cost, the agent terminates.

\remarkbox{\label{rmk:single_planning}Planning and framework optimizations target two economic axes. \emph{Friction cost compression}: GoT's thought merging synthesizes insights from independent branches into a single result---analogous to \emph{economies of scope}~\cite{varian1992micro}---while SWE-agent's interface reduces per-interaction overhead. \emph{Factor productivity enhancement}: search methods (ToT, LATS) diversify the exploration portfolio, and Voyager's skill library creates an experience curve in which accumulated experience lowers future average cost. Stop-loss budgeting balances these compounding returns against token expenditure.}


\clearpage
\section{Token Economics in Multi-Agent Systems}
\label{sec:multi_agent}

\chapterepigraph
  {Many production processes consist of a series of tasks, mistakes in any of which can dramatically reduce the product's value.}
  {Michael Kremer}
  {The O-Ring Theory of Economic Development, Q. J. Econ., 1993, p.~551.}

To deconstruct the token economics of MAS, this section proceeds from theoretical modeling to multi-dimensional optimization.
\Cref{subsec:multi_problem} formalizes the trade-off between specialization dividends and internal transaction costs.
\Cref{subsec:mas_benchmark} establishes the empirical foundation via measurement and benchmarking frameworks. Building on this, the remaining sections categorize state-of-the-art optimizations into two microeconomic paradigms.
First, Mechanism Design mitigates agency costs through extensive-margin agent orchestration in \Cref{subsec:mas_orchestration} and intensive-margin communication optimization in \Cref{subsec:mas_communication}. Second, Systemic Infrastructure eliminates physical transaction costs by enhancing computation efficiency in \Cref{subsec:mas_compute} and optimizing memory and knowledge coordination in \Cref{subsec:mas_memory}.

\subsection{Problem Modeling: Collaborative Token Economics}
\label{subsec:multi_problem}

\leadparagraph{Definition.}
Building upon the Coasian firm boundary model established in~\Cref{sec:phase2}, multi-agent token economics must be formalized as a communication topology optimization problem. Rather than merely tuning a scalar parameter, orchestrating a MAS involves optimizing a computational graph $G(V, E)$, where $V$ represents the set of agents (with scale $N = |V|$), and $E$ denotes their interaction pathways. This graph must balance the division of labor against coordination friction by managing two structurally distinct token categories: (1) Node Production Tokens ($M_{prod}$): These tokens are allocated to individual agents for specialized reasoning, with their total expenditure defining the Node Production Cost ($C_{prod}$). As $N$ increases, the reasoning burden on individual nodes is alleviated by specialization dividends, thereby compressing $C_{prod}$. (2) Consumed along the graph edges ($E$) for state synchronization and context passing, these tokens constitute the {Internal Transaction Cost ($C_T(G)$). As the network topology $G$ thickens, $C_T(G)$ scales super-linearly (e.g., $\mathcal{O}(|V|^2)$), reflecting rising coordination friction.

The total economic expenditure for a specific MAS topology $G$ is defined as:
\begin{equation}
    TC(G) = P_k \cdot K + w \cdot L + \underbrace{\sum\nolimits_{v\in V}\tilde{P}_{prod}\cdot M_{prod,v}}_{C_{prod}} + \underbrace{\tilde{P}_{comm} \cdot (M_{comm} + M_{waste})}_{C_T(G)}
\end{equation}

To achieve economic rationality, the MAS orchestration navigates a Pareto frontier over the graph space:
\begin{equation}
    \min_{K, L, M_{prod}, G} TC \quad \text{s.t.} \quad Y \ge Z
\end{equation}

\begin{figure}[htbp]
  \centering
  \includegraphics[width=0.9\linewidth, height=0.35\textheight, keepaspectratio]{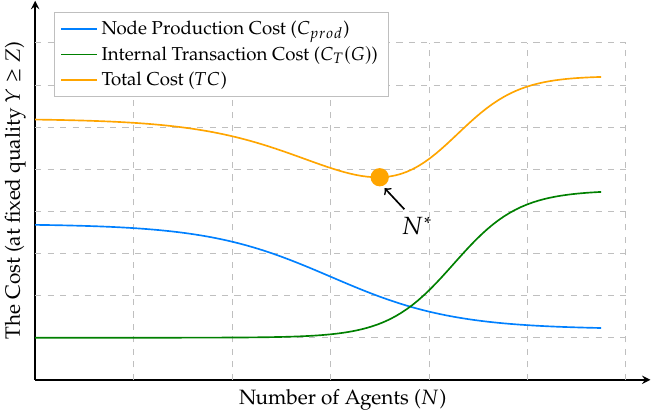}
  \caption{Organizational boundary optimization in MAS under a fixed-quality constraint ($Y \ge Z$). As the number of agents ($N$) increases, the Node Production Cost decreases due to specialization dividends. Conversely, Internal Transaction Costs grow super-linearly due to communication frictions. The Coasian optimum ($N^*$) occurs at the minimum of the Total Cost curve, representing the Pareto frontier resource allocation for the system.}
  \label{fig:multi_agent}
\end{figure}

\leadparagraph{Example.}
Consider a multi-agent system tasked with automatically producing a cross-domain report (\Cref{fig:multi_agent}). Initially, the system comprises a few generalist agents performing the entire workflow-from retrieval to formatting-resulting in high node production costs ($C_{prod}$) and limited output quality ($Y$). Introducing specialized roles (e.g., Retriever, Extractor) increases the agent count ($N$), yielding specialization dividends that significantly reduce $C_{prod}$ and enhance performance. However, once the scale exceeds the Coasian boundary ($N^*$), marginal gains are eclipsed by superlinear internal transaction costs ($C_T(G)$). These overheads, arising from role descriptions, state synchronization, and conflict resolution, manifest as a ``communication tax'' that can reach $\mathcal{O}(|V|^2)$ in dense topologies, driving the total cost ($TC$) back up. Consequently, while the bottleneck for $N < N^*$ is insufficient specialization, coordination friction dominates the system overhead when $N > N^*$. $N^*$ thus defines the most token-economical organizational boundary for a given task complexity.  

Before deploying structural interventions, \Cref{subsec:mas_benchmark} establishes the critical empirical foundation for cost attribution through measurement and benchmarking. Building upon this foundation, contemporary efforts to steer MAS configurations toward the optimal topology $G^*$ fall into two macro-paradigms (see \Cref{sec:concepts_te}):

\begin{itemize}[leftmargin=*, itemsep=4pt, topsep=4pt]
    \item \textbf{Paradigm A: \textsc{Engineering Optimization}}. Explored in \Cref{subsec:mas_compute} and \Cref{subsec:mas_memory}, this approach physically alters the diagram by eliminating transaction costs at the infrastructure level (e.g., cross-agent KV cache sharing and global memory coordination), effectively compressing the super-linear cost curve downward.
    \item \textbf{Paradigm B: \textsc{Resource Optimization}}. Detailed in \Cref{subsec:mas_orchestration} and \Cref{subsec:mas_communication}, this approach actively routes the system's execution path toward $G^*$ by pruning redundant communication topologies (extensive margin) and aligning local agent interaction protocols to curb agency costs (intensive margin).
\end{itemize}

\Cref{tab:mas_eco_mapping} summarizes how subsequent optimization strategies leverage these two paradigms to approach the Pareto frontier of collaborative token economics.

\begin{sidewaystable}[!htbp]
\caption{Multi-agent: technical solutions and economic mapping. (Paradigm A: \textsc{Engineering Optimization}; B: \textsc{Resource Allocation})}
\label{tab:mas_eco_mapping}
\centering
 
\renewcommand{\arraystretch}{1.2} 
 
\resizebox{\linewidth}{!}{
\begin{tabular}{@{} 
    >{\raggedright\arraybackslash}p{3.8cm} 
    >{\raggedright\arraybackslash}p{4.2cm} 
    >{\raggedright\arraybackslash}p{9.4cm} 
    >{\raggedright\arraybackslash}p{9.4cm} 
    c
    c
    @{}}
\toprule
\multirow{2}{*}{\textbf{Chapter}} & \multirow{2}{*}{\textbf{Core Objective}} & \multirow{2}{*}{\textbf{Representative Technical Solutions}} & \multirow{2}{*}{\textbf{Economic Mapping}} & \multicolumn{2}{c}{\textbf{Paradigm (\Cref{sec:concepts_te})}} \\
\cmidrule(l){5-6}
& & & & \textbf{A} & \textbf{B} \\
\midrule
\multirow{4}{=}{Communication Graph Pruning and Agent Elimination \newline (\Cref{subsec:mas_orchestration})} & 
\multirow{4}{=}{Extensive-margin Optimization: Adjust the number of agents, communication topology, and model selection to eliminate unnecessary token generation at the structural level.} & 
Communication Graph Pruning and Agent Elimination~\cite{zhang2024agentprune, wang2025agentdropout} & 
Locate the Coasian firm boundary ($N^*$); directly curtail super-linear internal transaction costs ($C_T(G)$) by pruning structurally redundant edges and nodes. & \xmark & \checkmark \\
\cmidrule(l){3-6}
 
& & 
Learned Topology Generation~\cite{zhang2024gdesigner, li2025argdesigner,jiang2025gtd} & 
Dynamically adjust organizational structures to balance specialization dividends against redundant agency costs ($M_{waste}$). & \xmark & \checkmark\\
\cmidrule(l){3-6}
 
& & 
Selective Participation and Cross-modal Debate Compression~\cite{zeng2025s2mad,wu2026debateocr} & 
Mitigate communication friction ($M_{comm}$) and coordination latency via selective participation and modality transformation. & \xmark & \checkmark \\
\cmidrule(l){3-6}
 
& & 
System-Level Routing and Budget-Aware Coordination~\cite{yue2025masrouter,zhou2025mass,jin2025corl} & 
Implement incentive-compatible routing, prioritize individual capability optimization over bureaucratic scaling ($N$), and enforce hard budget constraints that invoke expensive experts only when marginal contribution justifies cost. & \xmark & \checkmark \\
\midrule
 
\multirow{2}{=}{Agent Communication and Interaction Optimization \newline (\Cref{subsec:mas_communication})} & 
\multirow{2}{=}{Intensive-margin Optimization: Increase the information density of each message and optimize the content and format of cross-agent communication.} & 
Message-Level Communication Compression~\cite{chen2025optima,yang2025codeagents} & 
Increase inter-agent token information density through learned protocols and zero-cost formatting; alleviate principal-agent information asymmetry and reduce syntactic friction under bounded context windows. & \xmark & \checkmark \\
\cmidrule(l){3-6}
 
& & 
Runtime Resource Allocation and Quality Control~\cite{gandhi2024budgetmlagent,supervisoragent2025} & 
Lower the effective shadow cost of internal processing through cost-aware model substitution and inline quality control, truncating error propagation as a stop-loss mechanism against compounding agency costs ($M_{waste}$). & \xmark & \checkmark \\
\midrule
 
\multirow{3}{=}{Computation Efficiency \newline (\Cref{subsec:mas_compute})} & 
\multirow{3}{=}{Capital Efficiency Optimization: Reduce the actual processing cost per token through cache reuse at the underlying architecture level and cross-model sharing.} & 
Cross-context KV Cache Reuse~\cite{ye2025kvcomm,bian2026tokendance} & Exploit economies of scope in physical infrastructure to drive down the rental price of computational capital ($P_k$) and decouple latency from agent scaling. & \checkmark & \xmark \\
\cmidrule(l){3-6}
 
& & 
Representation-level Communication (Direct Vector Transmission)~\cite{kriuk2025qkvcomm} & Bypass the text tokenization bottleneck to reduce semantic redundancy and lower the physical component of internal transaction costs ($C_T(G)$). & \checkmark & \xmark \\
\cmidrule(l){3-6}
 
& & 
Multi-adapter and Cross-model Cache Sharing~\cite{jeon2026lragent,liu2024droidspeak} & 
Maximize capital productivity and structural TFP ($A$) in concurrent serving, sustaining specialization dividends under tight memory constraints. & \checkmark & \xmark \\
\midrule
 
\multirow{3}{=}{Memory Architecture and Retrieval Efficiency \newline (\Cref{subsec:mas_memory})} & \multirow{3}{=}{Knowledge Supply Chain Management: Balance information completeness and retrieval cost, and precisely control the usage of a limited context window.} & Memory Topology Design and Latent-Space Coordination~\cite{wu2025memory,srmt2024,legomem2024,zou2025latentmas,yu2026multi} & Transform recurrent inter-agent state synchronization overhead into scalable, shared cognitive capital; formalize consistency protocols analogous to cache-coherence models in distributed systems. & \checkmark & \xmark \\
\cmidrule(l){3-6}
 
& & 
Role-Specific and Self-Organizing Memory~\cite{yuen2025intrinsic,evocf2026} & Optimize the multi-agent knowledge supply chain by eliminating negative informational externalities ($M_{waste}$) through role-scoped templates and structured constraint induction. & \xmark & \checkmark \\
\cmidrule(l){3-6}
 
& & 
Token-Budget-Aware Retrieval and Capacity Control~\cite{rcr2025,gmemory2025,agentnet2025} & Operationalize the carrying-cost constraint via retrieval gating, structural compression, and capacity-controlled eviction policies, closing the feedback loop between the read path and the write path. & \xmark & \checkmark \\
 
\bottomrule
\end{tabular}
}
\end{sidewaystable}

\subsection{Measurement and Benchmarking of Token Consumption}
\label{subsec:mas_benchmark}

Optimizing MAS token economics requires fine-grained cost attribution across diverse roles, topologies, and pipeline stages. Recent work has begun establishing the empirical foundations for this emerging discipline.

\leadparagraph{Token Distribution Analysis and Taxonomies.}
\label{subsubsec:mas_taxonomy}AgentTaxo~\cite{wang2025agenttaxo} unifies agent roles into three archetypes (Planner/Reasoner/Verifier) and benchmarks token distributions across linear (MetaGPT~\cite{hong2024metagpt}), flat (\allowbreak CAMEL~\cite{li2023camel}), and hierarchical (AgentVerse~\cite{chen2023agentverse}) topologies, formalizing the notion of a ``communication tax'' and finding input-to-output ratios of 2:1--3:1 that identify context loading---not generation---as the dominant cost. Tokenomics~\cite{salim2026tokenomics} corroborates this asymmetry in software engineering, reporting that iterative code review consumes 59.4\% of ChatDev tokens, far exceeding initial generation. Bai et al.~\cite{bai2026howdo} extend this analysis to eight frontier LLMs across 500 SWE-bench tasks with four independent runs each, confirming that agentic coding consumes over 1000$\times$ more tokens than single-turn reasoning, with input tokens dominating at a ratio exceeding 150:1. Their study further demonstrates that token usage is inherently stochastic (up to 30$\times$ variance across runs of the same task), that accuracy peaks at intermediate cost levels before saturating, and that models vary substantially in token efficiency even on identical tasks. Frontier models also fail to predict their own token consumption before execution (Pearson $r \leq 0.39$), systematically underestimating actual costs, which highlights a fundamental gap between perceived and realized computational effort. MultiAgentBench~\cite{zhu2025multiagentbench} benchmarks four topologies (star, chain, tree, graph) with milestone-based KPIs, showing that graph topologies best balance performance against coordination overhead and that additional agents exhibit clear diminishing marginal returns.

\leadparagraph{Scaling Laws and Cross-Framework Evaluation.}
\label{subsubsec:mas_scaling}A comprehensive cross-framework evaluation~\cite{yin2025comprehensive} reveals a large gap between \textit{nominal} and \textit{effective} token costs under prompt caching ($\sim$\$0.07/M tokens), establishing system-level caching as a high-leverage intervention. A systematic 180-configuration study~\cite{kim2025towards} further identifies three empirical regularities: a \textit{tool--coordination tradeoff} (context-window crowding as $N$ grows), a \textit{capability ceiling} (coordination becomes net-negative when single-agent performance is already high), and \textit{architecture-dependent error amplification} in the absence of verification gates.

\remarkbox{\label{rmk:def_scaling}Measurement and benchmarking serve the same foundational role for MAS token economics that national income accounting serves for macroeconomic policy: without standardized cost attribution across agent roles, pipeline stages, and communication topologies, optimization lacks both a baseline and a target. Token distribution analysis and scaling law identification convert MAS cost reduction from ad hoc heuristics into a systematic practice with reproducible metrics, comparable baselines, and falsifiable hypotheses about where marginal effort yields the highest return.}

\subsection{Agent Orchestration and Scheduling}
\label{subsec:mas_orchestration}

Agent orchestration and scheduling optimizations target the ``organizational architecture'' of MAS. These optimizations adjust agent count, communication topology, role assignment, and model selection to reduce unnecessary token production at its structural source.
In economic terms, these methods operate on the \textit{extensive margin} of token production: rather than making each token more efficient, they eliminate entire categories of token expenditure by restructuring \textit{who participates}, \textit{how they are connected}, and \textit{what resources they command}.
The analogy to organizational economics is direct: just as firms reduce costs by eliminating redundant departments, consolidating communication channels, and matching employee skill levels to task difficulty, MAS orchestration methods seek token savings through structural redesign of the agent workforce.

\leadparagraph{Communication Graph Pruning and Agent Elimination.}
\label{subsubsec:mas_pruning}AgentPrune~\cite{zhang2024agentprune} models MAS as a spatial-temporal graph and applies low-rank-guided one-shot pruning of redundant messages, simultaneously improving robustness under adversarial attack, which suggests that redundancy is actively harmful rather than merely wasteful.
AgentDropout~\cite{wang2025agentdropout} extends this idea from edges to agent nodes by learning per-round degree scores and selectively removing low-contribution agents across different communication rounds, yielding significant reductions in both prompt and completion tokens.
The two approaches are complementary: AgentPrune optimizes ``\textit{what} is communicated,'' AgentDropout optimizes ``\textit{who} communicates.''

\leadparagraph{Learned Topology Generation.}
\label{subsubsec:mas_topology}A more ambitious family learns topologies from scratch rather than pruning from templates.
G-Designer~\cite{zhang2024gdesigner} formulates the multi-agent communication protocol as variational graph optimization, using a VGAE with sparsity and anchor regularization to balance efficiency against structural coherence.
ARG-Designer~\cite{li2025argdesigner} reframes topology construction as autoregressive graph generation, building agents and links incrementally from an empty graph; a metric-learning module supports extensible role pools, and a learned \textsc{END} token allows the model to terminate generation once a sufficient team is assembled, thereby avoiding oversized configurations.
GTD~\cite{jiang2025gtd} recasts topology design as conditional discrete graph diffusion, coupling a Graph-Transformer-based generator with a GAT-based surrogate reward model and zeroth-order guidance to navigate the accuracy-token Pareto frontier at finer granularity than single-shot methods.

\leadparagraph{Debate Efficiency and Selective Participation.}
\label{subsubsec:mas_debate}Multi-agent debate improves reasoning quality but incurs super-linear token costs as both agent count and round count grow.
S$^2$-MAD~\cite{zeng2025s2mad} introduces a decision-making mechanism based on viewpoint similarity that filters redundant exchanges and allows agents to skip rounds in which they have nothing novel to contribute, achieving substantial token savings with minimal accuracy loss.
DebateOCR~\cite{wu2026debateocr} takes a cross-modal approach: each round's textual debate history is rendered as an image and encoded via a SAM-CLIP vision pipeline into a compact set of vision tokens, reducing context growth from quadratic to linear in both agents and rounds while preserving, and in some cases improving, accuracy by suppressing stylistic noise.

\leadparagraph{System-Level Routing and Budget-Aware Coordination.}
\label{subsubsec:mas_routing}MasRouter~\cite{yue2025masrouter} unifies collaboration-mode selection, role allocation, and LLM routing in a single cascaded controller network, progressively constructing a MAS that balances effectiveness and efficiency for each query.
MASS~\cite{zhou2025mass} enforces the principle ``optimize individuals before structures'' via three-stage interleaved optimization: block-level prompt warm-up, workflow topology search within an influence-weighted design space, and workflow-level prompt refinement, recognizing that weak prompts cannot be remedied by stacking more agents.
CoRL~\cite{jin2025corl} implements centralized delegation through reinforcement learning with a multiplicative reward that zeroes out any budget overrun, teaching a lightweight controller to invoke expensive experts only when the marginal contribution justifies the cost and enabling controllable behavior across different budget regimes at inference time.

\remarkbox{\label{rmk:def_system_level}The orchestration results echo a central insight of Coase's theory of the firm~\cite{coase1937nature}: the optimal boundary of an organization is reached when the marginal cost of coordinating one additional internal unit equals the marginal benefit of its contribution.
In MAS, each added agent introduces coordination tokens---role descriptions, synchronization messages, conflict resolution---that constitute an internal ``transaction cost.''
The substantial token waste revealed by pruning and topology optimization is the multi-agent analog of bureaucratic overhead in an over-expanded firm: organizational structure is often a more powerful cost lever than individual worker productivity, and right-sizing the agent workforce yields savings inaccessible to prompt- or model-level tuning alone.
}

\subsection{Agent Communication and Interaction Optimization}
\label{subsec:mas_communication}

While orchestration methods determine \textit{who} communicates, communication optimization methods improve \textit{how} agents communicate.
This dimension operates on the \textit{intensive margin} of token production: each message is made more informative per token through training-driven protocol compression, format redesign, cost-aware model cascading, and runtime supervision.
The optimization target shifts from the structure of the agent graph to the content and efficiency of the messages flowing through it.

\leadparagraph{Message-Level Communication Compression.}
\label{subsubsec:mas_msg_compression}
Just as domain experts develop jargon that compresses lengthy explanations into terse phrases, agents can learn, or be reformatted into, more compact communication protocols.
Optima~\cite{chen2025optima} pursues this direction through training: it adopts an iterative pipeline that combines supervised fine-tuning with direct preference optimization on MCTS-diversified preference data, jointly optimizing task performance, token efficiency, and readability of inter-agent messages.
A substantial portion of communication redundancy can also be removed without any training.
CodeAgents~\cite{yang2025codeagents} replaces natural-language system prompts and plans with YAML role specifications and Python-style pseudocode, in which typed variables, control structures, and inline assertions encode planning and tool invocation in a more compact form.
The fact that this purely structural reformatting yields consistent improvements in both accuracy and token usage indicates that a non-trivial fraction of redundancy in untrained agent communication originates from the rhetorical and syntactic overhead of natural prose.
Taken together, these two lines of work bracket the compression spectrum: structured formatting realizes the readily available gains at zero training cost, whereas learned protocols extend further by adapting the communication code itself.

\leadparagraph{Runtime Resource Allocation.}
\label{subsubsec:mas_runtime_alloc}
Complementary to compressing messages, a second line of work reduces cost by routing queries to appropriately priced models and by intercepting errors before they propagate.
BudgetMLAgent~\cite{gandhi2024budgetmlagent} pairs a low-cost base model with an LLM cascade and a bounded ``ask-the-expert'' lifeline that caps invocations of more expensive models, demonstrating that strategic escalation of stronger models for a small fraction of steps can preserve, or even improve, task success rates while sharply lowering monetary cost.
SupervisorAgent~\cite{supervisoragent2025} operates at an even lighter weight, augmenting existing MAS frameworks with an LLM-free, rule- and embedding-based filter that triggers targeted interventions only at high-risk steps such as tool errors, repetitive loops, and excessively long observations, thereby reducing token consumption while maintaining or improving accuracy.
The shared insight is that early and cheap interception, whether through model selection or context filtering, prevents the compound accumulation of downstream repair costs and functions as in-line quality control rather than terminal inspection.

\remarkbox{\label{rmk:def_supervision}If orchestration optimizations adjust the \textit{quantity} of productive inputs, communication optimizations raise their \textit{total factor productivity} (TFP), namely the efficiency with which a given bundle of tokens is converted into task-relevant output.
This TFP gain operates through three channels that mirror the standard growth-accounting decomposition: raising the information content per token (technological progress), eliminating redundant or erroneous exchanges (waste reduction), and allocating expensive compute to the tasks where its marginal product is highest (allocative efficiency).
The compression ratios reported by both learned protocols and structural reformatting suggest that untrained inter-agent communication operates well inside its production-possibility frontier.
}

\subsection{Computation Efficiency}
\label{subsec:mas_compute}
Computation efficiency optimizations target the inference engine underlying MAS, reducing the \textit{per-token processing cost} through KV cache reuse, compression, and cross-model sharing.
While these methods do not directly reduce the number of tokens generated or transmitted, they lower the effective economic cost per token by eliminating redundant computation across agents, thus improving the capital efficiency of the inference infrastructure.
In the token production function, these methods reduce the cost of the ``compute capital'' input while holding the token quantity constant.

\leadparagraph{Cross-Context KV Cache Reuse.}
\label{subsubsec:mas_kvcache}
In multi-agent pipelines, identical text segments yield divergent KV caches when preceded by different agent-specific prefixes, a phenomenon termed the \textit{offset variance problem}.
KVComm~\cite{ye2025kvcomm} addresses this through a training-free anchor pool that estimates per-token cache offsets by interpolating from previously observed deviations, enabling substantial cache reuse across agents with negligible accuracy loss and order-of-magnitude reductions in prefill latency.
Q-KVComm~\cite{kriuk2025qkvcomm} takes a complementary approach by transmitting \textit{compressed KV representations} directly between agents via adaptive layer-wise quantization and heterogeneous model calibration, demonstrating that raw text is an unnecessarily expensive inter-agent medium when internal representations can be shared at high compression ratios while preserving semantic fidelity.
TokenDance~\cite{bian2026tokendance} targets the \textit{All-Gather} pattern common in synchronized MAS through three mechanisms: a round-aware prompt interface, collective KV Cache reuse (amortizing RoPE rotation, importance-based position selection, and selective recomputation once per round across all agents), and diff-aware block-sparse storage against a master copy. Together, these enable significant increases in agent concurrency and per-agent storage savings without additional accuracy degradation.

\leadparagraph{Multi-Adapter and Cross-Model Cache Sharing.}
\label{subsubsec:mas_multimodel}
When agents share a pretrained backbone but differ through lightweight adapters, further sharing opportunities arise.
LRAgent~\cite{jeon2026lragent} decomposes multi-LoRA value caches into a shared base component from the pretrained weights and compact low-rank per-adapter residuals, reconstructed on demand via a custom Flash-LoRA-Attention kernel that avoids materializing full-dimensional adapter contributions. This decomposition achieves throughput close to fully shared caching while preserving role-specific agent behavior.
DroidSpeak~\cite{liu2024droidspeak} extends cache sharing to different fine-tunes of the same foundation model by profiling layer-wise sensitivity and selectively recomputing only the critical layers, reusing KV and embedding caches for the remainder. This yields substantial prefill speedups with minimal quality degradation across diverse model pairs and tasks.

\remarkbox{\label{rmk:def_multi_adapter}
Computation efficiency methods achieve the multi-agent analog of \textit{capital deepening} in growth economics: holding the GPU capital stock fixed, KV cache reuse and compression increase the effective compute available per agent, shifting the marginal cost curve of token production downward.
The economic logic mirrors that of shared infrastructure in manufacturing: just as a common power grid lowers per-factory energy costs by amortizing fixed generation capacity, cross-context cache sharing amortizes the fixed cost of prefill computation across agents whose contexts overlap substantially.
The result is higher \textit{capital productivity}: more useful inference output is extracted from each unit of hardware investment, expanding the feasible scale of MAS deployment under a fixed infrastructure budget.
}

\subsection{Memory Architecture and Retrieval Efficiency}
\label{subsec:mas_memory}

While computation-efficiency techniques (Section~\ref{subsec:mas_compute}) focus on reducing the per-token processing overhead of context that has already entered the inference pipeline, memory management addresses a more fundamental question: \textit{which} context should be incorporated in the first place, and to what extent. Accordingly, memory management and knowledge-coordination mechanisms govern how MAS store, retrieve, update, propagate, and prune historical information across the agent collective.
In long-horizon multi-agent interactions, each memory retrieval injects
tokens into agent context windows, and each memory update generates tokens
for storage.
The design of the memory system directly shapes the token economics of
sustained multi-agent collaboration, creating a fundamental tension between
\textit{carrying cost}---every token of stored context injected into a
finite window displaces capacity for new reasoning---and
\textit{stockout cost}---missing a critical piece of historical information
degrades decision quality downstream.

\leadparagraph{Memory Topologies.}
\label{subsubsec:mas_memory_arch}%
A dedicated MAS-memory survey~\cite{wu2025memory} identifies three
canonical topologies with distinct economic properties:
\textit{agent-local} memory (efficient but silo-prone),
\textit{shared pools} (fast knowledge transfer but susceptible to a
``tragedy of the commons'' in which agents pollute the shared pool with
low-relevance information), and \textit{hybrid} designs with access
control, alongside open challenges such as coordinated forgetting,
sublinear scaling, and adaptive write policies.
Hybrid architectures address the carrying-versus-stockout tradeoff
directly: SRMT~\cite{srmt2024} couples each agent's personal memory vector
with a shared recurrent pool via cross-attention, while
LEGOMem~\cite{legomem2024} assigns full task memories to the orchestrator
and subtask-scoped memories to executors, containing the carrying cost of
shared context to the roles that actually require it.
LatentMAS~\cite{zou2025latentmas} pushes the shared-pool paradigm further
by dispensing with text-based exchange entirely: agents communicate via
layer-wise KV-cache transfers in continuous latent space, bypassing the
encoding--decoding cycle that dominates token cost in conventional
text-mediated collaboration and achieving 70--84\% token reduction relative
to text-based MAS while maintaining or improving task accuracy.
From a systems perspective, Yu et al.~\cite{yu2026multi} argue that the
absence of formal \textit{consistency models}---analogous to cache
coherence in multiprocessors---is the most critical gap, with access
granularity (the memory analog of cache line size) being a decisive yet
underexplored design parameter.

\leadparagraph{Role-Specific and Self-Organizing Memory.}
\label{subsubsec:mas_hetero_memory}%
Beyond inter-agent topology, the internal structure of each agent's memory store also shapes token efficiency.
Intrinsic Memory Agents~\cite{yuen2025intrinsic} equip each agent with a
structured, role-specific JSON template updated directly from its own
outputs (no separate summarizer call), substantially outperforming prior
multi-agent memory methods on PDDL planning while maintaining the highest
token efficiency---internalizing the negative externality that uniform memory imposes on agents whose roles do not require that information.
EvoCF~\cite{evocf2026} extends the self-organizing idea to embodied
multi-agent planning by maintaining typed memory records annotated with
preconditions, effects, and failure codes, from which symbolic constraints
are continuously induced and retrieved via compositional queries to guide
counterfactual plan generation---demonstrating that structured memory can
serve not only as a retrieval store but as an evolving rule library that
actively shapes plan search.

\leadparagraph{Token-Budget-Aware Retrieval and Capacity Control.}
\label{subsubsec:mas_memory_budget}%
Structured memory architectures reduce carrying cost by curating
\textit{what} is stored; a complementary line of work enforces explicit
token budgets on \textit{how much} is retrieved per step.
RCR-Router~\cite{rcr2025} maintains a shared interaction history and
routes context through three sequential gates---an Importance Scorer, a
Semantic Filter, and a Token Budget Allocator---that together minimize
redundant context replay while respecting a hard per-round token ceiling,
directly operationalizing the carrying-cost constraint as a first-class
system parameter rather than an implicit design goal.
G-Memory~\cite{gmemory2025} addresses the same constraint through structural compression: a three-tier graph hierarchy of insight, query, and interaction nodes enables bi-directional traversal that retrieves high-level generalizable insights alongside condensed interaction trajectories, replacing verbatim history replay with a lossy-but-compact summary that fits within tighter budgets.
On the write side, capacity control complements retrieval gating:
AgentNet~\cite{agentnet2025} maintains fixed-size per-agent memory modules
and prunes low-utility trajectories using composite signals of frequency,
recency, and uniqueness---the multi-agent analog of a least-recently-used
eviction policy that prevents unbounded accumulation of stale context.
Taken together, these methods instantiate the full inventory control loop
implied by the carrying-versus-stockout framing introduced above:
RCR-Router and G-Memory govern the \textit{reorder quantity} (how much to
retrieve), while AgentNet governs the \textit{warehouse capacity} (how much
to retain), closing the feedback cycle between the read path and the write
path.

\remarkbox{\label{rmk:def_memory}
Memory system design presents a classic inventory-theoretic tradeoff: carrying cost versus stockout cost.
The role-specific and self-organizing architectures surveyed above act as
just-in-time (JIT) delivery systems, supplying each agent only the information its role requires at the moment of consumption.
Capacity-control mechanisms (frequency- and recency-based eviction) serve
as the complementary \textit{disposal policy}, pruning memories whose
marginal retrieval value has fallen below their carrying cost.
The optimal memory policy, like the optimal reorder point, is not a
constant but shifts with task complexity, team size, and horizon length.}

\clearpage
\section{Token Economics of Intelligent Agent Ecosystems}
\label{sec:ecosystem}

\chapterepigraph
  {The firm has a role to play in the economic system if transactions can be organized within the firm at less cost than would be incurred if the same transactions were carried out through the market.}
  {Ronald H. Coase}
  {The Nature of the Firm: Meaning, Oxford: Oxford University Press, 1991, p.~48.}

This section extends the previous cost-minimization logic to shared LLM serving ecosystems. Once agents run on multi-tenant platforms, the core question becomes how scarce inference capacity is priced, routed, cached, and governed while required quality levels are met. \Cref{fig:eco_roadmap} provides a roadmap of the ecosystem-level analysis that follows.
\Cref{subsec:eco_problem} formalizes generalized-cost allocation under task-level quality, latency, and safety requirements. \Cref{subsec:eco_producer_consumer} examines producer-consumer interaction through dynamic pricing and prompt caching. \Cref{subsec:eco_producer_producer} addresses producer rivalry as open-weight models and ecosystem moats reshape token competition.
Next, \Cref{subsec:eco_regulator_market} examines how institutional rules convert carbon, access, and compliance externalities into token production costs.
Finally, \Cref{subsec:eco_dynamic} outlines the co-evolutionary cycle linking cost reductions, demand expansion, market restructuring, and regulatory responses.

\begin{figure*}[!htb]
  \centering
  \includegraphics[width=\linewidth]{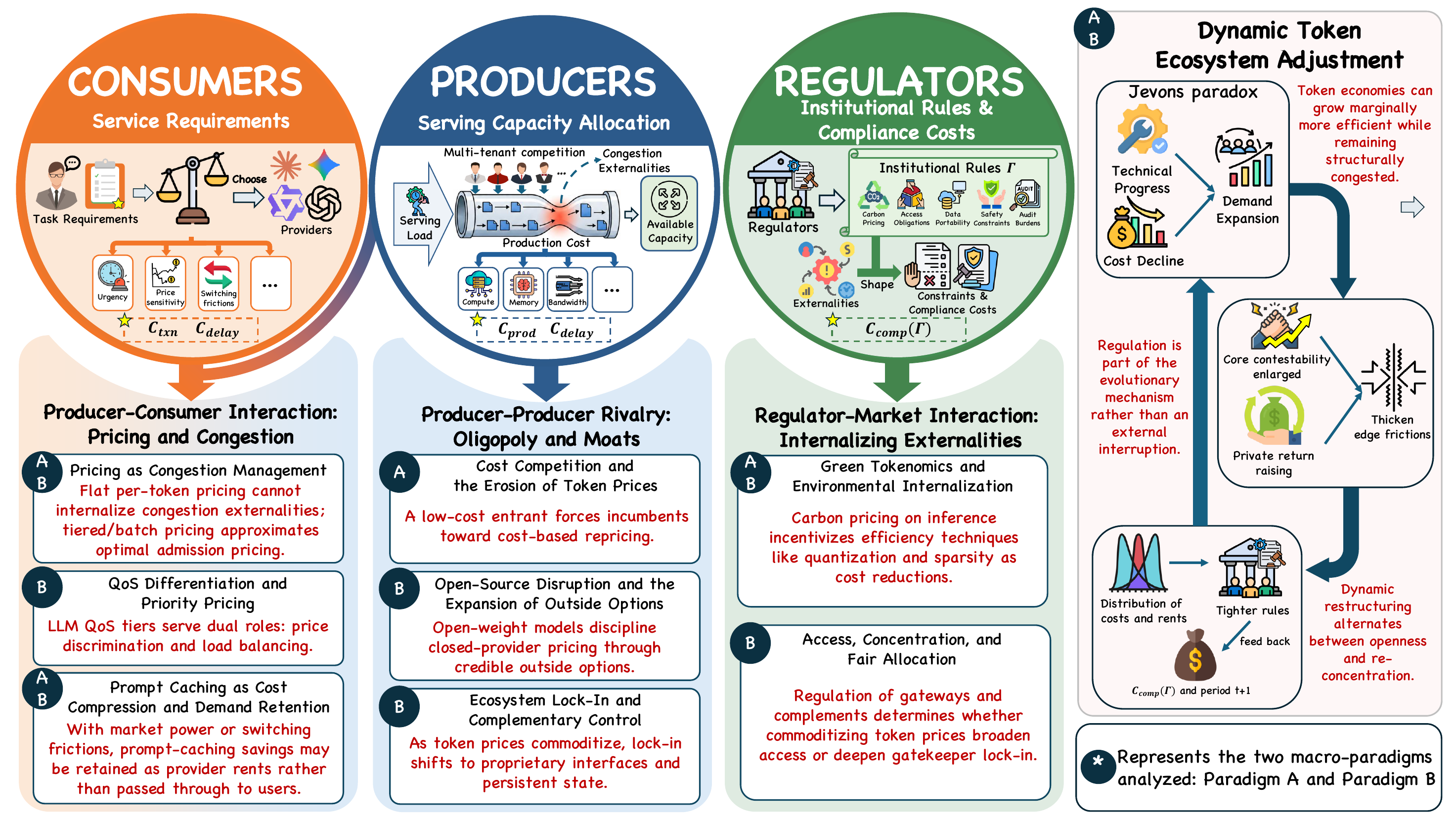}
  \caption{Roadmap of ecosystem-level token economics. The figure links shared serving infrastructure to producer-consumer interaction, producer-producer rivalry, regulator-market interaction, and dynamic ecosystem adjustment under constrained serving capacity.}
  \label{fig:eco_roadmap}
\end{figure*}

\subsection{Problem Modeling: Ecosystem Token Economics}
\label{subsec:eco_problem}

\leadparagraph{Definition.}
At the ecosystem level, token economics studies how users, agent workflows, model providers, tool-memory infrastructures, and open-weight outside options allocate scarce LLM serving capacity under institutional rules that shape costs, access, and compliance. Tokens serve as the common accounting unit because heterogeneous workflows compete for compute, memory, KV cache, bandwidth, low-latency slots, and compliance capacity. Building on \Cref{sec:concepts}--\Cref{sec:multi_agent}, we formalize the ecosystem problem as minimizing the cost of meeting required service levels:
\begin{equation}
    \min TC_E
    =
    C_{\mathrm{prod}}+C_{\mathrm{delay}}+C_{\mathrm{txn}}+C_{\mathrm{comp}}(\Gamma),
    \label{eq:eco_cost_benchmark}
\end{equation}
where $TC_E$ aggregates production, delay, transaction, and compliance costs. The minimization is evaluated under task-level performance and provider-level capacity constraints:
\begin{equation}
    Y_i\geq \bar{Y}_i,\quad
    S_i\geq \bar{S}_i,\quad
    \tau_i\leq \bar{\tau}_i\ \text{for latency-critical tasks},\quad
    \sum_{i\in\mathcal{R}_p} a_{i,p} \leq Cap_p,\quad \forall p\in\mathcal{P}.
    \label{eq:eco_constraints}
\end{equation}
Here, $Y_i$, $S_i$, and $\tau_i$ are ecosystem-level service outcomes. $Y_i$ denotes realized output quality. $S_i$ denotes safety or reliability performance. $\tau_i$ denotes realized latency. The barred terms are required thresholds set by benchmarks, application owners, service-level agreements, or regulators; the latency constraint applies when latency is task-critical. For provider $p\in\mathcal{P}$, $\mathcal{R}_p$ is the set of assigned requests, $a_{i,p}$ is the effective serving load of request $i$, and $Cap_p$ is the available inference capacity.

The four cost terms identify the main economic frictions in a shared inference ecosystem. $C_{\mathrm{prod}}$ is token-serving production cost inherited from the token production logic developed in~\Cref{sec:concepts}-\Cref{sec:multi_agent}, including compute, memory, KV-cache, bandwidth, and serving-stack efficiency. $C_{\mathrm{delay}}$ converts queueing latency, 
time-to-first-token, time-per-output-token, congestion, and SLA violations into generalized waiting cost. $C_{\mathrm{txn}}$ captures transaction and switching costs, including provider search, API adaptation, tool and protocol migration, cached-state loss, and memory-store portability. $C_{\mathrm{comp}}(\Gamma)$ captures compliance costs induced by institutional rules $\Gamma$, such as carbon pricing, access obligations, data portability requirements, safety constraints, and audit burdens.

These cost channels also organize actor roles: users enter through service requirements, urgency, price sensitivity, and switching frictions; providers through efficiency, capacity allocation, congestion, and lock-in; and regulators through rules that convert externalities into constraints and compliance costs.

\leadparagraph{Example.}
Consider a shared LLM platform serving a real-time IDE assistant, a nightly batch summarization job, and a compliance-sensitive enterprise retrieval workflow. These workloads draw on the same infrastructure but impose different quality, safety, latency, and governance requirements. The ecosystem problem is therefore to compare how pricing, routing, caching, interoperability, and compliance rules change the minimum cost of meeting required service levels.

Later subsections refine the aggregate terms in \Cref{eq:eco_cost_benchmark} through price and priority, cache-hit and state, provider efficiency, and policy variables. The remainder of this chapter analyzes ecosystem conflicts through two macro-paradigms (see \Cref{sec:concepts_te}):
\begin{itemize}[leftmargin=*, itemsep=4pt, topsep=4pt]
    \item \textbf{Paradigm A: \textsc{Engineering Optimization}} lowers the real serving cost of meeting a given quality target through MoE, quantization, prompt caching, and green-serving architectures.
    \item \textbf{Paradigm B: \textsc{Resource Allocation and Governance}} allocates scarce serving capacity under congestion, switching frictions, and institutional constraints through pricing, lock-in, interoperability, and compliance rules.
\end{itemize}
Finally, \Cref{subsec:eco_dynamic} synthesizes these paradigms into a co-evolutionary cycle of cost reduction, demand expansion, market power, and regulatory response.

\begin{sidewaystable}[!htbp]
\caption{Agent ecosystems: technical solutions and economic mapping. (Paradigm A: \textsc{Engineering Optimization}; B: \textsc{Resource Allocation})}
\label{tab:eco_eco_mapping}
\centering

\renewcommand{\arraystretch}{1.2} 

\resizebox{\linewidth}{!}{
\begin{tabular}{@{} 
    >{\raggedright\arraybackslash}p{3.8cm} 
    >{\raggedright\arraybackslash}p{4.7cm} 
    >{\raggedright\arraybackslash}p{8.5cm} 
    >{\raggedright\arraybackslash}p{9.8cm} 
    c
    c
    @{}}
\toprule
\multirow{2}{*}{\textbf{Chapter}} & \multirow{2}{*}{\textbf{Core Objective}} & \multirow{2}{*}{\textbf{Representative Technical Solutions}} & \multirow{2}{*}{\textbf{Economic Mapping}} & \multicolumn{2}{c}{\textbf{Paradigm (\Cref{sec:concepts_te})}} \\
\cmidrule(l){5-6}
& & & & \textbf{A} & \textbf{B} \\
\midrule
\multirow{3}{=}{Producer-Consumer Interaction \newline (\Cref{subsec:eco_producer_consumer})} & 
\multirow{3}{=}{Congestion Control and Service Differentiation: Manage queuing latency in multi-tenant shared infrastructure and maximize cache reuse.} & 
Scheduling Optimization~\cite{yu2022orca,agrawal2024sarathi,zhong2024distserve}  & 
Internalize congestion externalities via scheduling; optimize the wait-cost trade-off across heterogeneous latency SLAs. & \checkmark & \checkmark \\
\cmidrule(l){3-6}

& & 
Cache Reuse~\cite{kwon2023vllm,zheng2024sglang,qin2025mooncake} & 
Convert technical cache efficiency into customer switching costs; utilize platform-specific state as a retention asset. & \checkmark & \checkmark \\
\cmidrule(l){3-6}

& & 
Pricing Mechanisms: tiered QoS, batch processing discounts~\cite{li2023alpaserve,wu2023fastserve,yu2022orca} & 
Implement price-based admission control; use screening mechanisms to align user urgency with scarce serving capacity. & \xmark & \checkmark \\
\midrule

\multirow{3}{=}{Producer-Producer Rivalry \newline (\Cref{subsec:eco_producer_producer})} & 
\multirow{3}{=}{Cost Competition and Ecosystem Moat: Reduce the physical cost per token while building barriers to cross-platform migration.} & 
Cost Reduction Techniques~\cite{mixtral2024,deepseekv2,frantar2023optq,awq2024,leviathan2023speculative} & 
Cost asymmetry and intangible-infrastructure advantages discipline prices while preserving edge rents; efficiency gains shift the production frontier and trigger industry-wide repricing. & \checkmark & \xmark \\
\cmidrule(l){3-6}

& & 
Open-source Alternatives~\cite{touvron2023llama,deepseekv3} & 
Strengthen outside-option credibility and improve users' bargaining position to discipline closed-source monopoly rents. & \xmark & \checkmark \\
\cmidrule(l){3-6}

& & 
Protocol lock-in: proprietary tool-calling formats, MCP~\cite{anthropic2024mcp}& 
Balance protocol interoperability against ecosystem enclosure; managing rents via complementary interface assets. & \xmark & \checkmark \\
\midrule

\multirow{2}{=}{Regulator-Market Interaction \newline (\Cref{subsec:eco_regulator_market})} & 
\multirow{2}{=}{Compliance and Fair Allocation: internalize environmental burdens and govern access bottlenecks across the ecosystem.} & 
Green AI~\cite{you2023zeus,luccioni2024power} & 
Pigouvian correction for environmental impact; linking serving intensity to endogenous carbon-linked production penalties. & \checkmark & \checkmark \\
\cmidrule(l){3-6}

& & 
Interoperability and access governance~\cite{jeon2023compatibility,besley2023political}& 
Data portability, non-discrimination, and policy oversight determine whether interoperability reduces switching costs or reinforces dependence on dominant gateways. & \xmark & \checkmark \\
\midrule

Dynamic Ecosystem Adjustment \newline (\Cref{subsec:eco_dynamic}) & Long-term system evolution: a closed-loop feedback cycle of ``technology-driven cost reduction, demand surge, system congestion, and regulatory intervention''. & Cross-layer co-evolution: adaptive coordination across technology, markets, and policy~\cite{agrawal2024splitwise,zhong2024distserve,qin2025mooncake,zheng2024sglang} & Jevons paradox: reductions in inference cost do not eliminate compute scarcity. Instead, they induce a larger expansion in latent total demand, driving the system into a persistent feedback loop in which cost reductions stimulate demand, and demand in turn generates congestion. & \checkmark & \checkmark \\

\bottomrule
\end{tabular}
}
\end{sidewaystable}

\Cref{tab:eco_eco_mapping} summarizes how subsequent systemic interventions leverage these two paradigms to approach the Pareto frontier of ecosystem token economics.

\subsection{Producer-Consumer Interaction: Pricing and Congestion}
\label{subsec:eco_producer_consumer}

On shared inference platforms, the per-request latency at a given provider depends on that provider's utilization. Every admitted job raises load for others---the classical congestion externality~\cite{naor1969regulation,yang2020marginal,herzog2024city}---so price, priority, and latency are jointly determined.

Engineering choices shape per-request latency and provider utilization jointly. Orca~\cite{yu2022orca} uses iteration-level continuous batching to reduce idle GPU cycles, while Splitwise~\cite{agrawal2024splitwise}, DistServe~\cite{zhong2024distserve}, and Sarathi-Serve~\cite{agrawal2024sarathi} separate or chunk prefill so it does not crowd latency-sensitive decode traffic.

\leadparagraph{Pricing as Congestion Management.}
Commercial APIs approximate externality-adjusted admission pricing through batch discounts, provisioned-throughput classes, and queue-aware schedulers~\cite{fu2024efficient}, following congestion-pricing logic~\cite{naor1969regulation,afeche2004pricing,yang2020marginal,herzog2024city}: each admitted job is priced against its private service cost plus the marginal delay imposed on others. Flat per-token prices alone cannot express this externality because requests with the same token count can impose different waiting costs depending on timing, burstiness, cache state, and QoS tier.

\leadparagraph{QoS Differentiation and Priority Pricing.}
Priority pricing sorts users onto bundled price-delay menus without revealing private time valuations~\cite{jeon2022second}. In token markets, the QoS tier assigned to each request records its service class, screening delay-sensitive and delay-tolerant workloads across scarce serving capacity. Such menus act simultaneously as screening devices and load-balancing instruments: users reveal urgency through tier choice, while providers shift delay-tolerant demand away from congested real-time capacity.
In practice, QoS tiers are implemented through a combination of admission control and routing. OpenAI's Batch API offers a 50\% discount for delay-tolerant workloads (24-hour completion window), while synchronous endpoints carry no discount; Anthropic and Google offer provisioned-throughput tiers with guaranteed tokens-per-minute allocations at premium prices. At the infrastructure layer, service classes are enforced through GPU priority classes, SLA-tagged routers, and per-tier allocation on disaggregated prefill/decode clusters, making the QoS tier operationally concrete.

KV-cache hierarchies determine the cache-hit ratio. vLLM~\cite{kwon2023vllm} shares KV blocks across common prefixes, SGLang~\cite{zheng2024sglang} uses radix-tree longest-prefix matching with 2--5\(\times\) throughput gains on multi-turn workloads, and Mooncake~\cite{qin2025mooncake} pools KV blocks across nodes and routes requests to warm caches. On the pricing side, Anthropic charges cached input tokens at 10\% of the base input price (with a 25\% write surcharge on first occurrence), and OpenAI applies an automatic 50\% discount on repeated prefixes---making cache hits directly visible in the cost function while raising switching friction tied to provider-specific state.

\leadparagraph{Prompt Caching as Cost Compression and Demand Retention.}
Providers can pass cache-hit savings through to prices, retain them as margin, or reinvest them in quality. By analogy with cost pass-through and salience under imperfect competition~\cite{ganapati2020energy,kroft2024salience}, this split depends on market structure and discount visibility. Because cached state is tied to KV hierarchies and session memory, cache reuse can become a retention asset---raising user switching friction in proportion to the provider-specific state accumulated over a workflow~\cite{jeon2023compatibility}. At moderate-to-high switching costs, caching may therefore create more value by retaining workflows with accumulated provider-specific state than by immediate price pass-through.

\subsection{Producer-Producer Rivalry: Oligopoly and Moats}
\label{subsec:eco_producer_producer}

Producer rivalry runs along price, quality, latency, and switching friction rather than price alone~\cite{teh2023multihoming,tan2021effects,pellegrino2025product,jeon2023compatibility}. The three subsections trace cost compression, outside-option credibility, and complementary-asset control, producing commodity-like core pricing alongside persistent edge rents~\cite{de2024market,pellegrino2025product}.

MoE architectures, quantization, and speculative decoding lower provider-level unit serving costs. Mixtral and DeepSeek-V2 reduce active computation per token; GPTQ~\cite{frantar2023optq} and AWQ~\cite{awq2024} enable INT4 inference at minimal accuracy loss; speculative decoding~\cite{leviathan2023speculative} delivers 2--3\(\times\) wall-clock speedups. Stacked, these mechanisms can produce order-of-magnitude reductions in effective per-token cost.

\leadparagraph{Cost Competition and the Erosion of Token Prices.}
When DeepSeek-V2 substantially lowered input-token prices relative to GPT-4-class incumbents, the market price benchmark shifted downward. This reflects a cost-asymmetry mechanism: lower marginal serving costs allow efficient providers to discipline prices, while high fixed and intangible investments can preserve edge rents~\cite{de2024market,pellegrino2025product}. In the \Cref{subsec:eco_problem} framework, this channel operates through $C_{\mathrm{prod}}$.

Open-weight releases have made outside options credible and quantifiable. DeepSeek-V3~\cite{deepseekv3}, a 671B-total / 37B-active MoE model with freely available weights, achieved performance broadly comparable to GPT-4o at an API price roughly 1/20th of the incumbent; the Llama-3 family~\cite{touvron2023llama} demonstrated that open-weight 70B models match or exceed GPT-3.5, making self-hosted deployment on an 8${\times}$A100 node (roughly \$15K/month) economically viable at moderate scale. Collectively, these releases place an upper bound on the price--quality ratio that closed providers can sustain, even in the absence of user migration.

\leadparagraph{Open-Source Disruption and the Expansion of Outside Options.}
Industry reports that domestic Chinese API pricing has been ``anchored by DeepSeek'' since early 2025, even for users who never migrated. Open-weight models operate through outside-option credibility rather than realized substitution. Modern bargaining logic with dynamic outside options makes this channel precise: a credible release tightens the closed provider's participation constraint even at zero market share~\cite{mcclellan2024dynamic}.
This shifts the closed-provider price-quality frontier downward while leaving market structure nominally unchanged~\cite{de2024market}.

\leadparagraph{Ecosystem Lock-In and Complementary Control.}
As core token pricing commoditizes, rents migrate to complementary interface and state---the switching-friction asset~\cite{jeon2023compatibility,de2024market}. APIs, protocols, and persistent memory can sustain margins despite contestable core pricing.

Lock-in vectors are visible at both the interface and state layers. Proprietary tool-calling schemas still require adapter work across providers. MCP~\cite{anthropic2024mcp} lowers switching costs at the protocol layer, but host implementations keep model selection, authentication, and UX partially provider-specific, creating a standard-as-moat dynamic. Persistent memory services and hosted vector stores further raise export costs, so switching friction can grow even as core token prices fall.

\subsection{Regulator-Market Interaction: Internalizing Externalities}
\label{subsec:eco_regulator_market}

Regulation enters where private contracting leaves environmental or access externalities uninternalized~\cite{besley2023political,colmer2025does}. Security and privacy externalities are addressed separately in \Cref{sec:eco_security}.

\leadparagraph{Green Tokenomics and Environmental Internalization.}
Because per-token energy varies across models and tasks, unpriced serving choices can yield socially excessive emissions~\cite{luccioni2024power}. When the institutional environment $\Gamma$ includes a carbon price, the carbon-linked component of $C_{\mathrm{comp}}(\Gamma)$ is the Pigouvian correction~\cite{pigou1920welfare,colmer2025does,besley2023political}, targeted at serving intensity rather than training compute~\cite{strubell2019energy,patterson2022carbon}.

Quantization directly lowers the carbon-sensitive component of serving cost: empirical work reports that INT4 inference can reduce memory-bandwidth demand enough to cut per-token energy by roughly 40--75\%~\cite{xu2025resource}. Sparse MoE activation lowers active FLOPs per token, and speculative decoding compresses wall-clock time, while tools such as Zeus make these gains measurable at the job level~\cite{leviathan2023speculative,you2023zeus}. Once carbon is priced, providers with stronger quantization, sparsity, and scheduling discipline therefore face lower effective carbon cost per token.

\leadparagraph{Access, Concentration, and Fair Allocation.}
Access regulation targets routing gateways and complements through which dominant providers preserve power as token prices commoditize. Modern models of data portability and intangible-entry barriers imply that interoperability, non-discrimination, and portability rules determine whether falling serving costs broaden participation or deepen dependence on gatekeepers~\cite{jeon2023compatibility,de2024market,besley2023political}.

\subsection{Dynamic Token Ecosystem Adjustment}
\label{subsec:eco_dynamic}

\Cref{subsec:eco_producer_consumer}--\Cref{subsec:eco_regulator_market} treated production efficiency, institutional rules, provider-specific state, and demand scale as static. We now link them through feedback loops among efficiency, demand, market power, innovation, and regulation~\cite{casey2024energy,de2024market,besley2023political}.

This is consistent with a Jevons-like dynamic: cost declines from MoE, quantization, and serving optimization make new agentic workflows viable, which triggers further serving innovation such as prefill--decode disaggregation, distributed KV-cache pooling, and cache-aware routing~\cite{jevons1865coal,casey2024energy,agrawal2024splitwise,zhong2024distserve,qin2025mooncake,zheng2024sglang}.

\leadparagraph{Technical Progress, Cost Decline, and Demand Expansion.}
Large API price reductions alongside expanding aggregate usage are consistent with a Jevons-like rebound~\cite{jevons1865coal}. When demand is elastic, lower per-token cost can raise long-run use by admitting workloads across the participation threshold~\cite{casey2024energy}; token economies can become more efficient at the margin while remaining congested in aggregate.

\leadparagraph{Congestion, Competition, and Market Restructuring.}
The same period that saw core API prices decline also saw rapid MCP adoption and the proliferation of provider-specific memory services. Cheaper tokens enlarge core contestability while increasing returns to provider-specific state investments~\cite{jeon2023compatibility,de2024market}, so restructuring alternates between entry, multihoming, and re-concentration rather than converging monotonically~\cite{teh2023multihoming,tan2021effects,pellegrino2025product}.

\leadparagraph{Regulation as an Endogenous Response.}
The staged implementation of the EU AI Act, China's Interim Measures for Generative AI (2023), and scrutiny of AI energy footprints followed the scale effects of prior cost decline. This matches political-economy models in which regulation responds to costs, rents, externalities, and technological opportunities, then feeds back into \(C_{\mathrm{comp}}(\Gamma)\) and feasible cost structures for period \(t+1\)~\cite{besley2023political}.

Together, these loops show a token ecosystem in which efficiency, market power, and governance remain continuously renegotiated~\cite{jevons1865coal,casey2024energy,besley2023political}.

\clearpage
\section{A Security Perspective on Token Economics}
\label{sec:eco_security}

In agentic ecosystems, tokens are security-relevant whose risk characteristics can materially reshape marginal productivity and economic value. At least three channels matter for token economics. First, insecurity can reduce expected utility by degrading the reliability of retrieved context, generated outputs, and inter-agent communication. Second, it can raise the effective shadow price of tokens through filtering, provenance verification, access control, redundancy, and privacy-preserving computation. Third, it can generate non-local welfare losses when compromised tokens propagate through shared infrastructure and coordinated workflows.

To align the literature with this framework, we classify security risks along the token lifecycle (see~\Cref{tab:token_classification} and~\Cref{sec:concepts_classification}). Five categories are particularly salient: input-token risk, external-token risk, internal-token risk, inter-agent token risk, and market-level token risk.

\leadparagraph{Structural Overview.} The remainder of this section proceeds in four layers: \Cref{sec:security_taxonomy} classifies these five risk categories along the token lifecycle with a cross-referenced summary table; \Cref{sec:security_empirical} identifies the empirical channels through which security vulnerabilities reshape token economics; \Cref{sec:security_model} extends the token cost function to incorporate defense expenditures and attack loss expectations; and \Cref{sec:security_policy} concludes these findings into governance and institutional-design implications.

\begin{sidewaystable}[!htbp]
\caption{Security: representative literature aligned with the risk categories and empirical channels in \Cref{sec:eco_security}.}
\label{tab:security_refs}
\centering
\footnotesize
\renewcommand{\arraystretch}{1.2} 

\begin{tabularx}{\linewidth}{@{}
    >{\raggedright\arraybackslash}p{2.2cm}
    >{\raggedright\arraybackslash}p{1.8cm}
    >{\raggedright\arraybackslash}p{2.6cm}
    >{\raggedright\arraybackslash}X
    >{\raggedright\arraybackslash}X
    @{}}
\toprule
\textbf{Reference} & \textbf{Risk / Channel} & \textbf{Focus} & \textbf{Core Finding} & \textbf{Token-Economic Implication} \\
\midrule


Zou et al.~\cite{zou2023universal} & Input-Token Risk & Jailbreak / adversarial suffix & Aligned models remain vulnerable to universal and transferable attacks. & Input tokens can lower expected utility before any productive reasoning begins; defense adds recurring screening overhead. \\
\cmidrule(l){3-5}

Greshake et al.~\cite{greshake2023indirect} & Input-Token Risk & Indirect prompt injection & Untrusted external content can carry executable instructions into deployed LLM applications. & Input-side contamination propagates into downstream actions, raising verification cost before tokens can be treated as productive inputs. \\
\cmidrule(l){3-5}

Anil et al.~\cite{anil2024manyshot} & Input-Token Risk & Long-context jailbreaking & Larger context budgets can expand the attack surface together with capability. & The marginal value of additional context tokens is security-conditioned rather than monotonically increasing. \\
\cmidrule(l){3-5}

Zou et al.~\cite{zou2025poisonedrag} & External-Token Risk & Retrieval poisoning & Corrupted retrieved documents can substantially degrade retrieval-augmented generation. & External token acquisition carries a trust premium; provenance checks become part of the real token cost. \\
\cmidrule(l){3-5}

Hubinger et al.~\cite{hubinger2024sleeper} & Internal-Token Risk & Sleeper agents / deceptive alignment & Malicious behaviors may persist through safety training. & Security cost must include upstream auditing and deployment review, not only online filtering. \\
\cmidrule(l){3-5}

de Benedetti et al.~\cite{debenedetti2024agentdojo} & Inter-Agent Token Risk & Shared-memory / tool-chain prompt injection & Realistic agent environments expose prompt injection risks beyond static text benchmarks. & Security losses propagate across tool chains and memory, amplifying token waste in multi-step agent workflows. \\
\cmidrule(l){3-5}

Fang et al.~\cite{fang2024oneday} & Inter-Agent Token Risk & Tool-mediated exploit execution & Tool-enabled agents can exploit real one-day vulnerabilities. & Compromised tokens can become adversarial instructions with downstream action externalities, motivating stricter permissioning and rate limiting. \\
\cmidrule(l){3-5}

Liu et al.~\cite{liu2024prompt_injection} & Empirical Channel & Verification costs / attack-defense evaluation & Prompt injection can be formalized as a systematic attack-defense evaluation problem. & Security enters token economics as a measurable efficiency dimension rather than an anecdotal engineering concern. \\
\cmidrule(l){3-5}

Li et al.~\cite{li2022mpcformer} & Empirical Channel & Confidentiality overhead & Stronger confidentiality can be achieved, but with added communication and latency overhead. & Privacy protection raises the shadow price of each token through additional communication rounds and higher latency. \\
\bottomrule

\end{tabularx}
\end{sidewaystable}

\subsection{Risk Categories Along the Token Lifecycle}
\label{sec:security_taxonomy}

\leadparagraph{Input-Token Risk.} This risk arises when adversarial or malformed inputs enter the system. Jailbreak and prompt injection attacks are representative examples. Universal adversarial suffixes can reliably induce harmful behavior in aligned language models~\cite{zou2023universal}. Indirect prompt injection is particularly consequential in deployed systems: once models ingest untrusted external content, injected instructions could propagate to downstream applications~\cite{greshake2023indirect}. Long-context jailbreaking further suggests that larger context windows expand not only capability, but also attack surface~\cite{anil2024manyshot}.

\leadparagraph{External-Token Risk.} This risk emerges when tokens are sourced from untrusted external environments. PoisonedRAG shows that externally retrieved knowledge tokens may be corrupted before model ingestion~\cite{zou2025poisonedrag}. Although retrieval could be more cost-efficient than parametric storage, it also introduces exogenous quality uncertainty. External token acquisition, therefore, resembles the procurement of experience goods, whose quality cannot be fully ascertained ex ante.

\leadparagraph{Internal-Token Risk.} This risk concerns compromised model behavior that persists despite safety alignment. Sleeper Agents provides evidence that deceptive behaviors may remain latent after training~\cite{hubinger2024sleeper}. Accordingly, auditing, red teaming, and model evaluation should begin before deployment rather than only after failure.

\leadparagraph{Inter-Agent Token Risk.} Arises in settings where agents share context, memory pools, or routing infrastructures. Shared-memory poisoning and prompt propagation across agent workflows can turn local failures into system-wide cascades~\cite{debenedetti2024agentdojo}. Autonomous agents exploiting zero-day vulnerabilities further show how compromised tokens can become adversarial instructions that affect external services~\cite{fang2024oneday}.

\leadparagraph{Market-Level Token Risk.} This risk encompasses systemic disruptions to token markets. Denial-of-service attacks, capacity misreporting, and selective congestion may distort price signals and crowd out legitimate demand under adversarial load. These phenomena resemble artificial supply shocks that reduce aggregate welfare across the ecosystem \citep{dong2025an}.

These five categories are organized by the dominant stage at which insecurity enters or propagates along the token lifecycle. When a single attack spans multiple stages, we classify it by its primary intrusion or propagation channel for analytical clarity. Economically, this taxonomy implies that security risk acts as a token-level risk premium (Remark~\ref{rmk:security_risk_premium}).

\remarkbox{\label{rmk:security_risk_premium}Security changes the economic meaning of token volume. A nominally cheap token may become expensive once its provenance, integrity, and downstream actionability are uncertain. Input-token attacks, retrieval poisoning, latent model behaviors, inter-agent propagation, and market-level congestion therefore operate like a risk premium added to the baseline shadow price $\tilde{P}_{m,i}$. The relevant price is not only the API fee or latency cost, but the expected cost of transforming an untrusted token into a reliable, productive input. Under this view, secure token allocation becomes a problem of adverse-selection management: systems must distinguish high-quality information goods from contaminated ones before allowing them to enter the production function.
}

\subsection{Empirical Security Channels: Evidence and Mechanisms}
\label{sec:security_empirical}

The empirical literature identifies three principal mechanisms through which security vulnerabilities reshape token economics. These mechanisms operate through three parallel channels: they raise effective shadow prices, reduce expected net utility, and inflate latency or coordination overhead.

\leadparagraph{Verification Costs Alter the Shadow Price of Tokens.} Before retrieved content can be used as a productive input, systems must incur additional token and latency costs for provenance verification, redundancy, and trust calibration. Liu et al.~\cite{liu2024prompt_injection} formalize this as a benchmarkable attack--defense problem. Retrieval efficiency and retrieval trustworthiness are jointly determined; optimizing one without the other yields an incomplete economic account.

\leadparagraph{Agentic Actions Transform Tokens into High-Stakes Outputs.} 
When models transition from passive response generation to autonomous action execution, output tokens effectively become executable control signals capable of invoking tools, modifying files, and interacting with external services. Consequently, a compromised token incurs not only wasted computational resources but also potentially significant downstream consequences. Security mechanisms such as permission control, rate limiting, and sandbox isolation should therefore be treated as integral components of the token allocation and execution framework, rather than mere implementation details.

\leadparagraph{Confidentiality Constraints Increase Communication Overhead.} 
Privacy-preserving inference frameworks, such as \textit{MPCFormer}, demonstrate that stronger confidentiality guarantees inevitably introduce additional latency and communication overheads~\cite{li2022mpcformer}. Consequently, privacy constraints fundamentally alter the shadow price of tokens. Even for identical tasks, enforcing stronger privacy guarantees may require additional communication rounds, tighter synchronization, and substantially more expensive secure computation.

Collectively, these mechanisms imply that the security-adjusted net utility of tokens deviates substantially from their nominal value. The relevant marginal question is not whether additional tokens improve task performance per se, but whether they increase \emph{security-adjusted} net utility once price, harm, and coordination overhead are jointly considered.

\subsection{An Economic Cost Model Under Security Constraints}
\label{sec:security_model}

Building upon the token economics framework established in \Cref{sec:concepts}, we extend the cost function to incorporate security-related expenditures. We define the baseline shadow price of tokens as $\tilde{P}_{m,i} = P_m + w \cdot \tau_i$, where $P_m$ denotes the per-token procurement price, $w$ represents the opportunity cost of time for human participation, and $\tau_i$ denotes unit latency, as defined in \Cref{sec:concepts,sec:ecosystem}:
\begin{equation}
    \tilde{P}_{m,i} = P_m + w \cdot \tau_i,
    \label{eq:shadow_price}
\end{equation}

Security constraints introduce additional cost components, which we formalize as:
\begin{equation}
    C_{\mathrm{total}}(K,\pi)
    =
    C_{\mathrm{compute}}(K)
    +
    C_{\mathrm{coord}}(K)
    +
    C_{\mathrm{defense}}(\pi)
    +
    \mathbb{E}[L_{\mathrm{attack}} \mid K,\pi],
    \label{eq:security_cost}
\end{equation}
Here, $K$ denotes physical computational capital (e.g., GPU memory and FLOPS), as defined in \Cref{sec:concepts}. The term $\pi$ denotes the defense portfolio, including filtering, provenance verification, sandboxing, cryptographic protection, and redundant evaluation.

Within the broader token-economics framework, these terms enter through distinct channels. Defense requirements can raise the effective shadow price of tokens by adding verification latency and coordination overhead; attack risk lowers expected net utility through the loss term; and in multi-agent or ecosystem settings, security controls can enlarge coordination and compliance burdens. The joint budgeting of productive and defensive tokens therefore becomes part of the broader security-aware allocation problem discussed in \Cref{sec:future}.

The central economic insight is that $C_{\mathrm{defense}}(\pi)$ and $\mathbb{E}[L_{\mathrm{attack}} \mid K,\pi]$ are inversely related. Stronger defensive measures raise immediate expenditure while reducing expected downstream losses. The optimal level of security investment is determined by balancing these opposing forces, consistent with the congestion- and compliance-sensitive resource allocation logic discussed in \Cref{sec:ecosystem}.

This formulation also explains the non-linear nature of security losses at the ecosystem level. In interconnected agent systems, local failures may propagate through communication graphs, shared memory pools, toolchains, and routing layers. As a result, a seemingly minor prompt injection or retrieval poisoning event may deplete shared token budgets, trigger repeated verification cycles, and induce a system-wide welfare shock that substantially exceeds the magnitude of the initial failure.

\subsection{Policy Implications: Governance as Economic Infrastructure}
\label{sec:security_policy}

Under the security-adjusted framework developed above, several ecosystem-level phenomena admit a distinct economic interpretation. These interpretations collectively motivate the view of security governance as economic infrastructure (Remark~\ref{rmk:security_governance}):

\begin{itemize}
    \item \textbf{Security-Conditioned Token Utility:} External and communication tokens are economically valuable only to the extent that their provenance and trustworthiness can be verified. Once compromised, they cease to function as productive inputs and instead become channels for adversarial amplification, thereby lowering expected net utility even when nominal token volume remains unchanged. This parallels the adverse selection problem in information economics, wherein quality is imperfectly observable before consumption.

    \item \textbf{Higher Shadow Prices under Defense:} Input filtering, retrieval provenance verification, access control, and privacy-preserving inference increase the effective shadow price of tokens through additional latency, verification overhead, and communication costs, as formalized in \Cref{eq:shadow_price}.

    \item \textbf{Market Failure and Strategic Manipulation:} Denial-of-service attacks, capacity misreporting, and selective congestion can distort market-clearing conditions and reduce aggregate welfare under adversarial load. These phenomena resemble supply shocks in commodity markets and impose negative externalities on all participants reliant on token-based infrastructure~\cite{varian1992micro,zhang2025crabs}.

    \item \textbf{Governance as Institutional Infrastructure:} Mechanisms such as reputation systems, reserve capacity, provenance auditing, and zero-trust orchestration sustain the efficiency of token markets. These should be understood not merely as security interventions but as institutional arrangements that mitigate the aforementioned market failures through lower coordination, verification, and market-wide externality costs.
\end{itemize}

\remarkbox{\label{rmk:security_governance}Security mechanisms play the same role in agentic token production that quality control and market institutions play in industrial economies. Filtering, provenance checks, sandboxing, redundant evaluation, access control, and reserve capacity may appear to add overhead, but they intercept defective or adversarial tokens before private failures become system-wide losses. This creates a prevention--loss tradeoff: stronger defenses raise $C_{\mathrm{defense}}(\pi)$, while weak defenses increase $\mathbb{E}[L_{\mathrm{attack}} \mid K,\pi]$ through retries, repair tokens, privacy leakage, tool misuse, and cascading coordination failures. Provenance auditing, zero-trust orchestration, reputation systems, and priority queues should therefore be understood as institutional infrastructure that internalizes security externalities and preserves token-market efficiency under adversarial load.
}

In sum, these channels show that security is not an external compliance layer but a first-order determinant of token cost, utility, and welfare. Accordingly, token economics in agentic systems cannot be modeled as a frictionless allocation problem. Security constraints reshape the feasible frontier by altering expected utility, marginal costs, and systemic welfare. This argument is consistent with Paradigm C in \Cref{sec:concepts_te} and directly motivates the research agenda on security overhead and security-aware token budgeting developed in \Cref{sec:future}.

\clearpage
\section{Trends and Opportunities}
\label{sec:future}

The formalization of Token Economics remains in its infancy. While this survey has established a rigorous dual-view framework spanning single-agent optimization, multi-agent coordination, and ecosystem-level market dynamics, numerous critical frontiers demand sustained intellectual investment. We distill these into eleven interconnected research directions. To provide a structured roadmap for the community, we organize this section into two complementary parts: \Cref{subsection:future-trends} outlines six major trends that reflect the ongoing paradigm shifts in agent inference, memory utilization, and security overhead; \Cref{subsection:future-opportunities} identifies five Emerging Opportunities that highlight the next generation of theoretical and infrastructural challenges, ranging from differentiable token budgeting to dynamic real-time markets.

\begin{figure*}[!htbp]
  \centering
  \includegraphics[width=\linewidth]{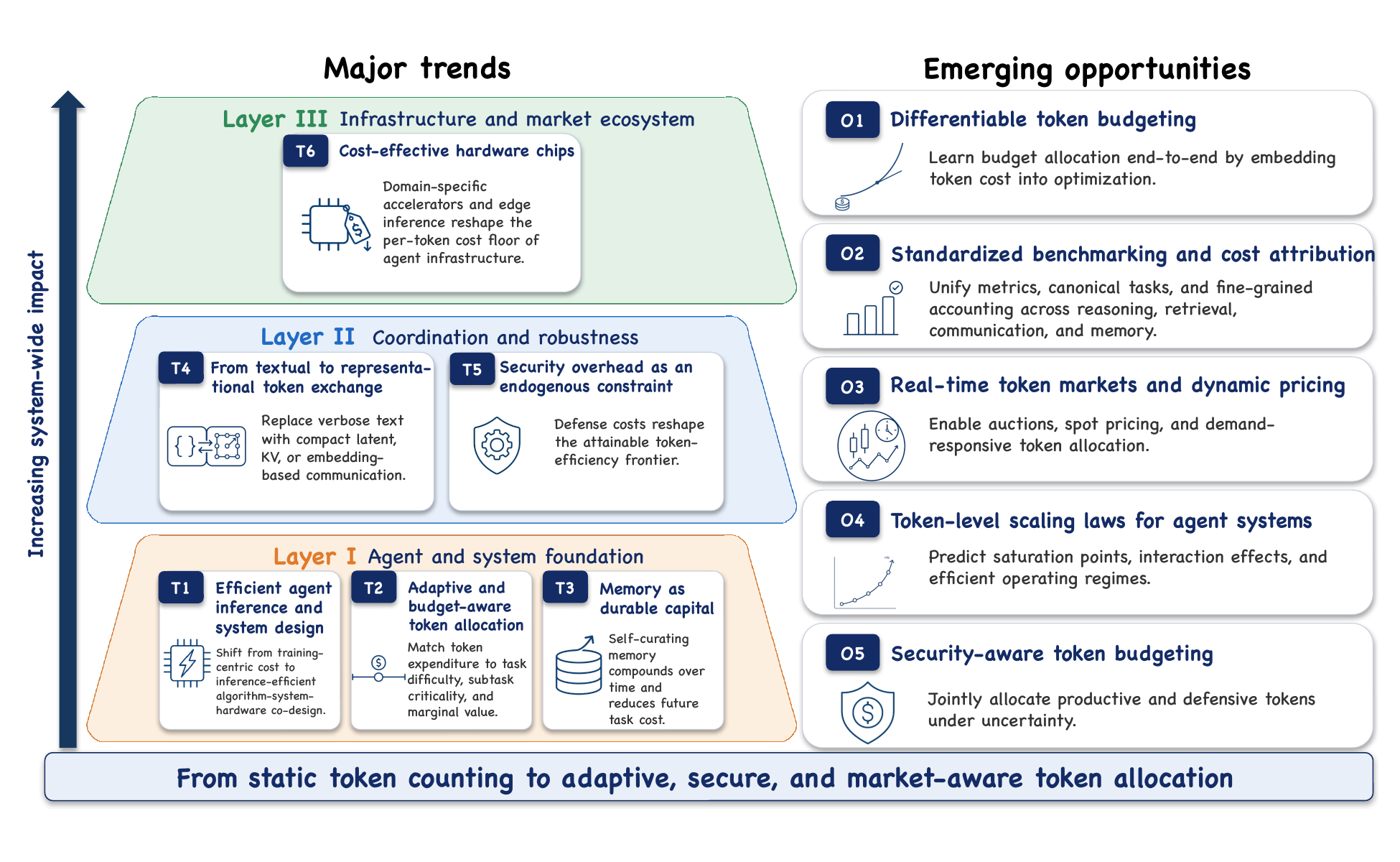}
  \caption{Trends and opportunities in token economics, organized into three ascending layers of system-wide impact (left) and five emerging research opportunities (right).}
  \label{fig:future-roadmap}
\end{figure*}

\subsection{Major Trends in Token Economics}
\label{subsection:future-trends}
\labeledlead{T1}{Efficient Agent Inference and System Design.}
As token demand continues to grow, the economic burden associated with tokens is increasingly shifting from the one-off, compute-intensive training stage to the persistent and distributed inference stage. Accordingly, greater emphasis is being placed on how to more effectively leverage pretrained models to solve downstream tasks, particularly with the support of context engineering and harness engineering~\cite{he2026harness}. Looking forward, future research may focus on efficient agent-centric inference and system design, including acceleration stacks across the algorithm, system, and hardware layers to better support agent workloads.

\labeledlead{T2}{Adaptive and Budget-Aware Token Allocation.}
A second trend is the shift from static, uniform token expenditure toward dynamic, budget-conscious allocation that matches token investment to task difficulty and subtask criticality.
Early exit mechanisms~\cite{yang2026dynamicearlyexit,jiang2025flashthink} terminate reasoning when marginal returns fall below marginal costs.
Adaptive-RAG~\cite{jeong2024adaptive} routes queries to different retrieval intensities based on complexity.
BudgetMLAgent~\cite{gandhi2024budgetmlagent} cascades from cheap to expensive models on a per-step basis, and CoRL~\cite{jin2025corl} trains controllers with hard budget constraints via multiplicative reward signals.
The unifying principle is the internalization of economic reasoning into the agent's decision loop: rather than treating tokens as a free resource to be consumed liberally, modern systems increasingly model the cost--benefit tradeoff of each additional token explicitly---implementing, in effect, the marginal analysis that has long been the cornerstone of microeconomic theory.

\labeledlead{T3}{Memory as Durable Capital with Compounding Returns.}
Agent memory systems are evolving from passive storage into active, self-curating knowledge assets that exhibit increasing returns to experience.
Generative Agents' Reflection mechanism~\cite{park2023generative} synthesizes higher-order abstractions that improve all subsequent decisions.
Reflexion~\cite{shinn2023reflexion} converts task failures into reusable episodic capital.
Voyager's skill library~\cite{wang2023voyager} transforms first-time exploration costs into near-zero-cost reusable assets.
In multi-agent settings, A-MEM~\cite{xu2025amem} maintains a self-organizing knowledge graph that grows denser rather than simply larger.
The trend is toward memory systems that function as \emph{appreciating assets}---each unit of token investment in memory construction yields dividends across an expanding horizon of future tasks, creating a learning curve where cumulative experience progressively reduces per-task costs.

\labeledlead{T4}{From Textual to Representational Token Exchange.}
A pronounced trend across both single-agent and multi-agent settings is the migration of information exchange from the surface level of natural language text to the deeper level of continuous representations.
In single-agent systems, latent reasoning~\cite{hao2025coconut,amos2026thinkingstates} replaces verbose chain-of-thought traces with compact hidden-state computation, bypassing the tokenization bottleneck entirely.
In multi-agent systems, Q-KVComm~\cite{kriuk2025qkvcomm} transmits compressed KV cache representations between agents instead of re-encoding text, while DebateOCR~\cite{wu2026debateocr} converts textual debate histories into fixed-dimensional visual embeddings.
These developments converge on a common insight: natural language, despite being the native medium of LLMs, is a surprisingly \emph{lossy and expensive} communication format---rich in syntactic redundancy, rhetorical filler, and stylistic noise that inflate token counts without proportional information gain.
The emerging paradigm treats text as a human-facing interface layer and representation as the internal medium of computation and coordination, fundamentally decoupling "what the model thinks" from "how it communicates."

\labeledlead{T5}{Security Overhead as an Endogenous Efficiency Constraint.}
Prior analysis treats tokens as trusted inputs valued purely by information content. In practice, agentic systems make security an integral and costly part of the token lifecycle. Empirical studies of guardrails show a persistent tradeoff: stronger security reduces usability, with no jointly optimal point. Return-on-control evidence further indicates a large variance in defense effectiveness, implying that unguided investment leads to misallocation.
These costs extend system-wide. In multi-agent pipelines, failures propagate and amplify: prompt injection operates at the pipeline level, and local errors can converge into false consensus. Differentially private inference likewise introduces multiplicative overhead, requiring repeated executions across partitions.
Thus, security overhead has become an endogenous constraint shaping the token efficiency frontier. Ignoring defense costs will systematically overestimate achievable efficiency.

\labeledlead{T6}{More Cost-Effective Hardware Chips.}
As the economic importance of tokens becomes increasingly evident on a global scale, the cost-effectiveness of hardware chips is receiving growing attention. The focus is gradually shifting from the upper bound of hardware performance to how hardware clusters can be deployed efficiently at scale. To support this transition, low-cost, energy-efficient hardware with high token throughput at scale is likely to become an important trend in the future.

\subsection{Emerging Opportunities for Token Economics}
\label{subsection:future-opportunities}
The trends identified above, combined with the open challenges surfaced throughout our survey, reveal several promising research opportunities that we believe will define the next phase of token economics.

\labeledlead{O1}{Differentiable Token Budgeting.}
Current budget-aware systems rely on discrete mechanisms---hard loop limits, threshold-based cascading, or RL-trained controllers---to govern token allocation.
A natural next step is to make token budgeting \emph{end-to-end differentiable}: embedding cost signals directly into the model's loss function so that gradient-based optimization can learn to allocate tokens across reasoning steps, tool calls, and retrieval operations in a jointly optimal manner.
Early exit under adaptive stopping~\cite{yang2026dynamicearlyexit,jiang2025flashthink} and CoRL's multiplicative reward~\cite{jin2025corl} represent initial steps in this direction, but a fully differentiable framework that treats the token budget as a Lagrangian constraint during training remains an open and impactful research target.
Such a framework would enable models to internalize the price of computation as a first-class training signal, producing agents that are ``natively economical'' rather than externally constrained.

\labeledlead{O2}{Standardized Benchmarking and Cost Attribution.}
Despite the pioneering efforts of AgentTaxo~\cite{wang2025agenttaxo}, Tokenomics~\cite{salim2026tokenomics}, and MultiAgentBench~\cite{zhu2025multiagentbench}, the field lacks a unified benchmarking standard for token economics.
Existing evaluations differ in their cost accounting conventions (whether to count cached tokens, how to price input vs.\ output tokens, and whether to include failed attempts), making cross-study comparison difficult.
A standardized token economics benchmark suite---with agreed-upon cost metrics, canonical task sets spanning diverse complexity levels, and reproducible evaluation protocols---would accelerate progress by enabling fair comparison across methods and establishing community baselines.
Such a benchmark should incorporate not only aggregate token counts but also fine-grained attribution across functional categories (reasoning, communication, retrieval, memory, error correction), enabling researchers to identify and target the highest-leverage optimization surfaces.

\labeledlead{O3}{Real-Time Token Markets and Dynamic Pricing.}
Current token pricing is static: providers charge fixed rates per input and output token regardless of demand, task complexity, or time-of-day utilization patterns.
As agent systems grow in sophistication and scale, an opportunity emerges for \emph{dynamic token markets} where prices reflect real-time supply--demand conditions.
On the supply side, heterogeneous compute resources (GPU clusters of varying capability and utilization) could offer tokens at different price points.
On the demand side, agents with varying urgency and quality requirements could bid for token budgets accordingly.
Auction-based allocation mechanisms, spot pricing for off-peak inference, and futures contracts for guaranteed capacity are all natural extensions of the economic framework developed in this survey.
Such markets would enable Pareto-improving trades between cost-sensitive and latency-sensitive workloads, improving aggregate welfare across the token economy.

\labeledlead{O4}{Token-Level Scaling Laws for Agent Systems.}
Neural scaling laws~\cite{kaplan2020scaling,hoffmann2022training} have transformed our understanding of how model performance relates to parameter count, dataset size, and compute.
An analogous body of work is needed for \emph{agent-level} token scaling: How does task performance scale with total token expenditure across reasoning, communication, retrieval, and memory?
Preliminary findings---such as the diminishing and eventually negative returns to agent count observed in MultiAgentBench~\cite{zhu2025multiagentbench}, the capability ceiling effect~\cite{kim2025towards}, and the concavity of the reasoning investment spectrum (\S\ref{subsubsec:reasoning})---suggest that agent token scaling laws may be substantially more complex than model scaling laws, exhibiting phase transitions, interaction effects, and non-monotonic regimes.
Establishing rigorous, predictive scaling laws for agent token expenditure would provide practitioners with the theoretical foundation to size agent systems optimally, avoiding both under-investment (too few tokens for convergence) and over-investment (diminishing returns beyond saturation).

\labeledlead{O5}{Security-Aware Token Budgeting.}
Current work treats token budgeting and security separately: the former maximizes utility under a budget assuming homogeneous tokens, while the latter evaluates defenses without budget constraints. No framework jointly allocates \emph{productive tokens} (reasoning, retrieval, communication) and \emph{defensive tokens} (filtering, verification, sandboxing). The security cost model in this survey suggests a unified objective-minimizing total expected cost across compute, coordination, defense, and attack loss-analogous to optimal insurance design.
This problem is more complex than standard allocation. Attack distributions are unknown and non-stationary, requiring robust or Bayesian approaches. Defense controls exhibit interaction effects, yet are mostly evaluated in isolation. In multi-agent systems, defenses create positive externalities, leading to underinvestment under decentralization.
Solving this would integrate efficiency and security into a single Pareto frontier, enabling principled allocation of token budgets between utility and defense.

\clearpage
\section{Conclusion}
\label{sec:conclusion}
As LLM agents reshape the technological landscape, the token has transcended its role as a mere metric of computation to become the fundamental economic primitive of agentic AI. However, existing research on agentic inference remains highly fragmented---predominantly confined to low-level systems engineering without the guidance of formal economic theory. To bridge this gap, this survey established a Token Economics framework that unifies computational systems with economic theory and organizes the full lifecycle of token allocation into a coherent field-level blueprint.

Tracing the architectural evolution of LLM-agent systems, we organized this emerging area from theoretical foundations to single-agent optimization, multi-agent coordination, ecosystem-level allocation, and security economics. Across these layers, we showed how tokens can be understood as factors of production, media of exchange, and units of account, and how this perspective unifies questions of production, cost, communication overhead, and market friction within a single analytical language. At the micro-level, we modeled single-agent inference as a cost-minimization problem constrained by target output quality, achieved through dynamic factor substitution. At the meso-level, we showed how multi-agent collaboration incurs transaction and agency costs, and we reviewed structural interventions to mitigate these diseconomies of scale. At the macro-level, we analyzed agentic infrastructures as open, multi-tenant markets shaped by congestion, pricing, and mechanism design. Finally, we argued that security should not be treated as an external afterthought, but as an endogenous source of token-economic attrition that reshapes the efficiency frontier itself.

Research in Token Economics remains nascent. The shift from heuristic scheduling toward end-to-end Differentiable Token Budgeting, persistent memory capital with compounding returns, representational token exchange beyond natural-language redundancy, and dynamic token markets will define the next frontier for LLM agents. Ultimately, inference acceleration and algorithmic optimization are no longer purely engineering choices; they are economic propositions that determine whether agentic AI can achieve commercial viability, systemic robustness, and sustainable scalability. We hope this survey can serve not only as a roadmap of the current literature, but also as a common language and principled foundation for the community to design the next generation of robust, efficient, secure, and scalable agent systems.

\smallskip
\leadparagraph{Disclaimer.}
This survey represents our initial effort to systematically bridge the rapidly evolving fields of LLM agent architectures and microeconomic theory. As a first version, it may inevitably contain oversights, and we intend to continuously refine and update this manuscript. We warmly welcome constructive feedback, discussions, and corrections from the research community via email at \url{yuxichen@zju.edu.cn} or \url{lihuan.cs@zju.edu.cn}.


\clearpage
\bibliographystyle{unsrt}
\bibliography{main}

\end{document}